\documentclass[twoside]{article}

\usepackage[accepted]{aistats2025}

\usepackage[utf8]{inputenc}
\usepackage{amsmath,amsthm,amssymb,nicefrac,microtype}
\usepackage{graphicx,subfig,wrapfig}
\usepackage{booktabs,color,colortbl,multirow}
\usepackage{array,lipsum,pifont,tikz}
\usepackage{caption}
\usepackage{makecell}
\usepackage{comment}
\usepackage{float}

\usepackage{natbib}
\usepackage{hyperref}
\hypersetup{colorlinks,linkcolor=red!80!black,citecolor=blue!80!green,urlcolor=blue}
\usepackage[capitalize,nameinlink]{cleveref}
\crefname{section}{\S}{\S\S}
\Crefname{section}{\S}{\S\S}
\creflabelformat{equation}{#2\textup{#1}#3}

\usepackage[textsize=tiny]{todonotes}

\newcommand{\cmark}{\textcolor{green!40!black}{\ding{51}}}
\newcommand{\xmark}{\textcolor{red!60!black}{\ding{55}}}

\def\f{\mathbf{f}}
\def\x{\mathbf{x}}
\def\y{\mathbf{y}}

\def\e{\mathbf{e}}

\def\1{\mathbf{1}}
\def\0{\mathbf{0}}

\def\D{\mathcal{D}}

\def\N{\mathcal{N}}
\def\E{\mathbb{E}}

\def\be{\boldsymbol\epsilon}

\def\var{\operatorname{var}}

\usepackage{tcolorbox}
\definecolor{mylightgray}{gray}{0.95}
\newtcolorbox{mybox}{colback=mylightgray,colframe=mylightgray,top=1.2pt,bottom=1.2pt,right=1.8pt,left=1.8pt}

\begin{document}

\runningtitle{Robust Classification by Coupling Data Mollification with Label Smoothing}

\runningauthor{Markus Heinonen, Ba-Hien Tran, Michael Kampffmeyer, Maurizio Filippone}

\twocolumn[

\aistatstitle{Robust Classification by\\ Coupling Data Mollification with Label Smoothing}

\vspace{-3ex}

\aistatsauthor{
Markus Heinonen \\
    Department of Computer Science \\
    Aalto University, Finland \\
    \And 
  Ba-Hien Tran \\
  Mathematical and Algorithmic Sciences Lab \\
  Huawei Paris Research Center, France \\
  \AND
  
  \vspace{0ex}
  
  Michael Kampffmeyer \\ \hspace{-6ex}
  Department of Physics and Technology \\ \hspace{-6ex}
  UiT The Arctic University of Norway, Norway \\ \hspace{-6ex}
  \And \hspace{-0ex}
  Maurizio Filippone \\ \hspace{-0ex}
  Statistics Program \\ \hspace{-0ex}
  KAUST, Saudi Arabia \\ \hspace{-0ex}
  
}

]

\begin{abstract}
Introducing training-time augmentations is a key technique to enhance generalization and prepare deep neural networks against test-time corruptions. Inspired by the success of generative diffusion models, we propose a novel approach of coupling data mollification, in the form of image noising and blurring, with label smoothing to align predicted label confidences with image degradation. The method is simple to implement, introduces negligible overheads, and can be combined with existing augmentations. We demonstrate improved robustness and uncertainty quantification on the corrupted image benchmarks of CIFAR, TinyImageNet and ImageNet datasets.
\end{abstract}

\section{INTRODUCTION}
\label{sec:intro}

Image classification is a fundamental task in deep learning that has enjoyed a lot of developments throughout the last decade, while continuing to attract attention thanks to contributions in, e.g., tuning training protocols  \citep{hendrycks2019augmix,wightman2021resnet} and representation learning \citep{radford2021learning}.
Training-time \emph{augmentations}, whereby models are exposed to input variations to accelerate training ({CutMix}, {MixUp}) \citep{zhang2017mixup,yun2019cutmix}, mimic test time variability ({AugMix}, {TrivialAugment}) \citep{hendrycks2019augmix,muller2021trivialaugment}, or induce robustness against spurious signal ({PixMix}) \citep{hendrycks22}, also represent a rich family of works in this literature. 
These approaches result in improvements on test accuracy \citep{torchvision} and on corrupted out-of-distribution test images \citep{hendrycks2019benchmarking}. 
Somewhat surprisingly, augmentations typically assume that the labels are not degraded, even if an augmentation leads to object occlusion.  
In a parallel line of work on label smoothing, training labels are decreased to reduce network over-confidence \citep{szegedy2016rethinking,muller2019does} with calibration improvements \citep{guo2017calibration}.

\begin{figure*}[t]
    \centering
    \includegraphics[width=0.92\textwidth]{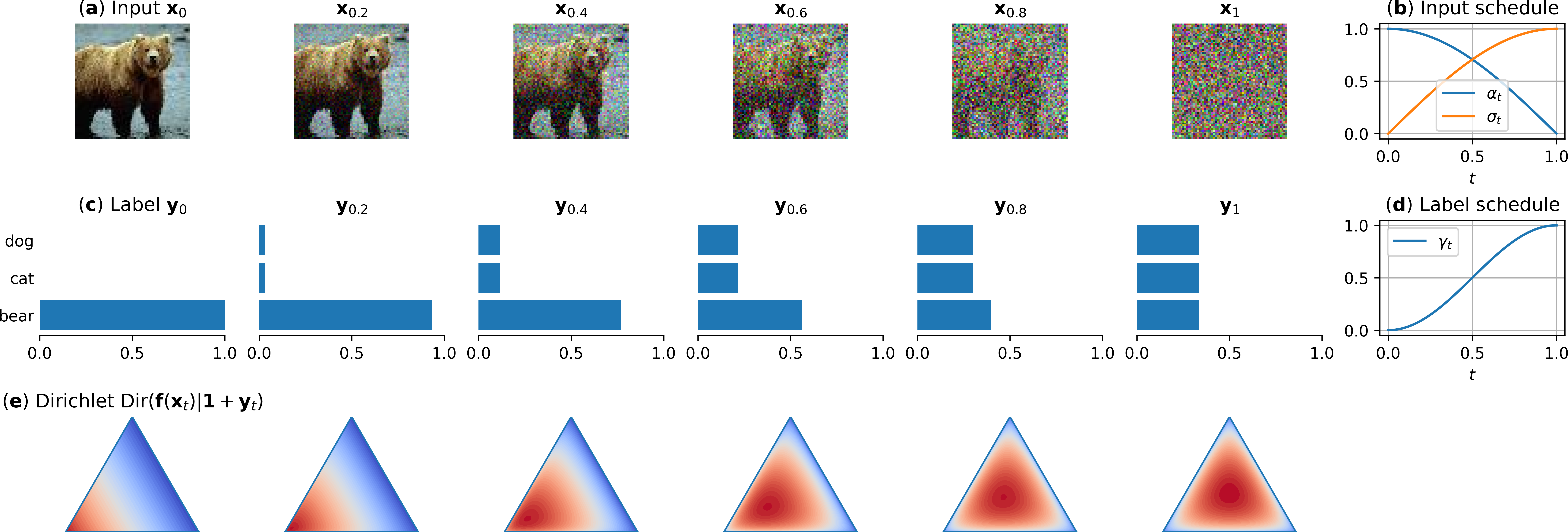}
    \caption{\textbf{Mollification augments training with perturbed images (a) and smoothed labels (c)}. Input mollification and label smoothing follow monotonic schedules (d), which reflect the signal-to-noise ratio of the images (b). Label smoothing prefers predictions whose distribution matches label uncertainty (e). The method extends to an arbitrary number of classes.}
    \label{fig:overview}
    \vspace{-2.5ex}
\end{figure*}

In this paper, we propose a novel approach to improve robustness of classification models to input corruptions by taking inspiration from the literature on Generative Diffusion Models (GDMs) \citep{song2020score}, which are currently dominating the state-of-the-art.
A key feature of GDMs is the mechanism of {\em data mollification} obtained by corrupting inputs by adding noise \citep{song2020score} or removing signal \citep{hoogeboom2022blurring}. 
Recently, \citet{tran2023} provided evidence that data mollification is responsible for dramatic improvements in density estimation and quality of generated samples in likelihood-based generative models.
In this work, we aim to leverage such a key component behind the success of GDMs to improve performance and robustness of classifiers. 
In our proposal, data mollification in classification materializes in the coupling of \emph{training-time input mollification}, in the form of noising and blurring, with \emph{training-time label mollification} in the form of label smoothing \citep{szegedy2016rethinking}. 
Label smoothing decreases label confidence based on the intensity of corruption of the inputs (see \cref{fig:overview}), which encourages the network to match the predicted label confidence to the amount of noise in the input image.

Our contributions are: 
\textbf{(1)} 
We present image classification by training under mollified inputs and labels, and discuss connections with Dirichlet distributions and tempering;
\textbf{(2)}
We provide a probabilistic view of data mollification and label smoothing, which is simple to implement and can be combined with existing augmentation techniques to improve performance on test-time corrupted images;
\textbf{(3)}
We demonstrate that our approach allows for a better calibration on in-distribution CIFAR-10/100, TinyImageNet
and ImageNet, while yielding strong performance on out-of-distribution data (e.g., sub-10\% error rate on CIFAR-10-C). %

\section{RELATED WORKS}
\label{sec:related}

\paragraph{Data Augmentation.}

In contrast to earlier works focusing on removing patches from images (see, e.g., \citet{devries2017cutout}), recent works provide evidence of improving performance of CNNs %
by mixing input images in various ways.
One of the earliest attempts is MixUp \citep{zhang2017mixup}, which fuses pairs of input images and corresponding labels via convex combinations. 
\citet{teney2024selective} and \citet{yao2022improving} improved out-of-distribution robustness of MixUp by selectively choosing pairs of inputs using a predefined criterion. 
Later, {CutMix} \citep{yun2019cutmix} %
copies and pastes patches of images onto other input images to create collage training inputs.
More recently, {AugMix} \citep{hendrycks2019augmix} proposed mixing multiple naturalistic transformations from {AutoAug} \citep{cubuk2019autoaugment}, while maintaining augmentation consistency. %
\citet{muller2021trivialaugment} further proposed {TrivialAug} %
where augmentation type and strength are sampled at random.

In other approaches, which we consider orthogonal to our work due to their complexity or focus on test-time augmentations, \cite{schneider2020improving} and \cite{zhang2022memo} adapt to test-time covariate shifts by modulating the statistics of batch-norm layers or by performing ensemble prediction of a test-image under all possible corruptions.
DeepAugment \citep{hendrycks2021many} uses an additional network to generate augmented data, and \citet{lee2020compounding}  involves training an ensemble. %
In contrast, our method can be directly applied to any architecture without modifications or extra networks. However, we believe that combining existing methods with our approach could yield results superior to each method individually.

\paragraph{Label Smoothing.}
Label smoothing \citep{szegedy2016rethinking} has been proposed to improve robustness of neural networks when employing data augmentation. 
For any input image, reusing the same label for the derived augmented images during training induces overconfidence in predictions. 
Label smoothing addresses this problem by mixing the one-hot label with a uniform distribution while still using the popular cross-entropy loss. 
This simple modification %
has been shown to yield better calibrated predictions and uncertainties \citep{muller2019does,thulasidasan2019calibration}.
The main question is how to optimally mix the one-hot encoded label with a distribution over the other classes \citep{Kirichenko2023understanding}, and recent works propose ways to do so by looking at the confidence in predictions over augmented data  \citep{maher21instance,quin2023autolabel} or relative positioning with respect to the classification decision boundary \citep{li20structured}.

\paragraph{Probabilistic perspectives.}
Perturbed inputs have been argued to yield degenerate \citep{izmailov21posterior} or tempered likelihoods \citep{kapoor2022uncertainty}. Several works study how label smoothing mitigates issues with the so-called cold-posterior effect \citep{wenzel20posterior,bachmann22tempering}. \citet{nabarro22augmentation} proposes an integral likelihood similar to our approach, and applies Jensen lower bound approximation. \citet{kapoor2022uncertainty} interprets augmentations and label degradation through a Dirichlet likelihood, and analyze the resulting biases. 
The work in \citet{wang23BA} considers neural networks with stochastic output layers, which allows them to cast data augmentation within a formulation involving auxiliary variables.
Interestingly, this yields a maximum-a-posteriori (MAP) objective in the form of a logarithm of an expectation, which is optimized via expectation maximization. 
\citet{lienen21labelrelax} propose label relaxation as an alternative to label smoothing by operating with upper bounds on probabilities of class labels. 
Recently, \citet{wu24} use {MixUp} data augmentation to draw samples from the martingale posterior \citep{fong2023martingale} of neural networks.

\section{ROBUST CLASSIFICATION THROUGH MOLLIFICATION}
\label{sec:methods}

We consider supervised machine learning problems with $N$ observed inputs $\x$ and labels $\y$ so that $\D = (\x_n,\y_n)_{n=1}^N$, where labels are one-hot-encoded vectors $\y \in [0,1]^C$ over $C$ classes. 
We focus on image classification problems, which pose a challenge due to their large dimensionality and the need for robustness.

We pose a question inspired by the success of Generative Diffusion Models \citep{ho2020denoising,song2020score}: 
\begin{mybox}
    \emph{Can we leverage the features that make GDMs so successful in generative modeling and density estimation to obtain classifiers $p(\y | \x)$ that perform well on in-distribution inputs while being robust on out-of-distribution inputs?}
\end{mybox}
In the GDM literature, tangential discussions around this question have arisen by either adding a classifier on top of the generative model, or carving out the predictive conditionals $p(\y|\x)$ from the generative joint distribution \citep{ho2022classifier,li2023diffusion}. Our goal, instead, is to exploit the mechanism of {\em data mollification} which characterizes GDMs for the predictive task directly, without building costly GDMs.

In GDMs, the data distribution $p_0(\x) := p(\x)$ is successively annealed into more noisy versions $p_t(\x)$ for $t \in [0,1]$ until $p_1(\x) \sim \N(0,I)$ is pure noise. 
In image classification, where we aim to estimate a decision boundary among classes, we can follow the same principle and assume a noisy image distribution $p_t(\x_t | \x)$. \begin{mybox}
{\em How should we choose the label distribution of $\y_t$ given the true label $\y$ and the noisy image $\x_t$?} 
\end{mybox}
In the extreme $\x_1 = \be$ the input is pure noise, and the original label surely loses its validity. 
Therefore, we argue that we need to monotonically degrade the label while noising the input. 
Similar considerations can be made for input mollification based on blurring.  

In this section, we formulate and analyze these concepts, and show that the application of input mollification and label smoothing to classification can be viewed through the lens of \emph{augmentations}, while in the experiments we demonstrate that this enables significant improvements in predictive performance under corruptions.

\subsection{Augmented likelihood}

The standard likelihood of a product of pointwise likelihoods stems from De Finetti's theorem \citep{cifarelli1996finetti} and from i.i.d. data sampling assumption $\{\x_n,\y_n\} \overset{\mathrm{iid}}{\sim} p(\x,\y)$,
\begin{align}
    \log p(\D | \theta) &= \sum_{n=1}^N \log p(\y_n | \x_n, \theta),
\end{align}
where $\theta$ denote model parameters.

We can introduce transformations $T_\phi(\x)$ whose parameters $\phi$ are auxiliary variables that are marginalized out as follows
\begin{equation} \label{eq:auglik}
    \mathcal{L} = \log p(\D|\theta) = %
    \sum_{n=1}^N \log \int p(\y_n | \x_n, \phi, \theta) p(\phi) d\phi,
\end{equation}
where we assume that the prior over the transformations is independent on inputs $\x_n$ and parameters $\theta$. %
The expression in \cref{eq:auglik} gives rise to two main questions:
\begin{mybox}
    \emph{How can we efficiently deal with the log-expectation in \cref{eq:auglik} to optimize $\theta$ under input mollification?}
\end{mybox}
Each likelihood term in \cref{eq:auglik} is an expectation under the distribution over transformations, and as such its log-sum form cannot be easily estimated unbiasedly using mini-batches of corrupted inputs.
A simple and effective solution is to use Jensen's inequality to obtain a lower bound:
\begin{equation} \label{eq:bound_auglik}
    \mathcal{L} \geq  
    \sum_{n=1}^N \int \log p(\y_n | \x_n, \phi, \theta) p(\phi) d\phi.
\end{equation}
Each integral can then be approximated in an unbiased fashion with Monte Carlo by sampling corruption intensities from $p(\phi)$, while the sum over $N$ can also be approximated unbiasedly through mini-batching; this leads to the possibility to obtain stochastic gradients to optimize $\theta$. 

The second question is now: 
\begin{mybox}
    \emph{What form should we choose for $p(\y_n | \x_n, \phi, \theta)$?}
\end{mybox}
Arguably, \emph{any} image transformation has a probability to degrade its associated label, since it is difficult to think that transformations do not affect label information at all; in an image, transformations such as pixel noise, crops, zooms or rotations can easily lead to the image losing some of its core meaning.
The likelihood $p(\y_n | \x_n, \phi, \theta)$ should be defined in a way that captures this behavior. 
We dedicate the rest of this section to an in-depth discussion of how to define a likelihood function for the label associated with corrupted inputs. 
Before doing so, it is useful to discuss mollification strategies for the inputs, which we do next. 

\begin{figure*}[t]
    \centering
    \includegraphics[width=0.87\textwidth]{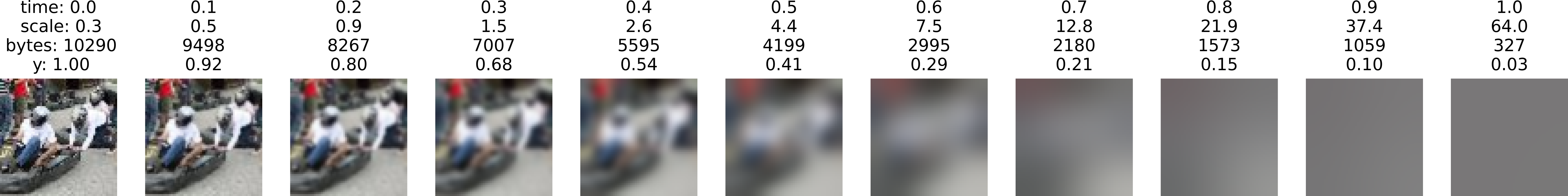}
    \caption{The logarithmic heat blurring schedule on a TinyImageNet 64$\times$64 image.}
    \label{fig:blur}
    \vspace{-1ex}
\end{figure*}

\subsection{Input Mollification}
The two key modalities of input mollification consist of drowning the image in Gaussian noise \citep{song2020score,ho2020denoising}, or applying low-pass filters to the image by blurring \citep{bansal2024cold}. By repurposing them as training-time augmentations, the model learns to ignore spurious noise patterns, while blur removes textures information and requires the model to perform classification under low-frequency representations of images \citep{rissanen2022generative}. 

\paragraph{Noising.} 
We follow a standard way of mollifying inputs by following the literature on GDMs: \citep{ho2020denoising}
\begin{align} \label{eq:input:diffusion}
    \x_t^\mathrm{noise} &= \underbrace{\cos(t \pi /2) }_{\alpha_t} \x + \underbrace{\sin(t \pi / 2)}_{\sigma_t} \be,
\end{align}
where we mix input image $\x$ and noise $\be \sim \N(0,I)$ in proportions of $\alpha_t$ and $\sigma_t$ according to the variance-preserving cosine schedule \citep{pmlr-v139-nichol21a}. We assume the \emph{temperature} parameter $t \in [0,1]$ in unit interval. We also assume image standardization to $\E[\x_n] = 0$ and $\operatorname{Var}[\x_n] = \mathbf{1}$.

\begin{figure}[t]
    \centering\includegraphics[width=0.33\textwidth]{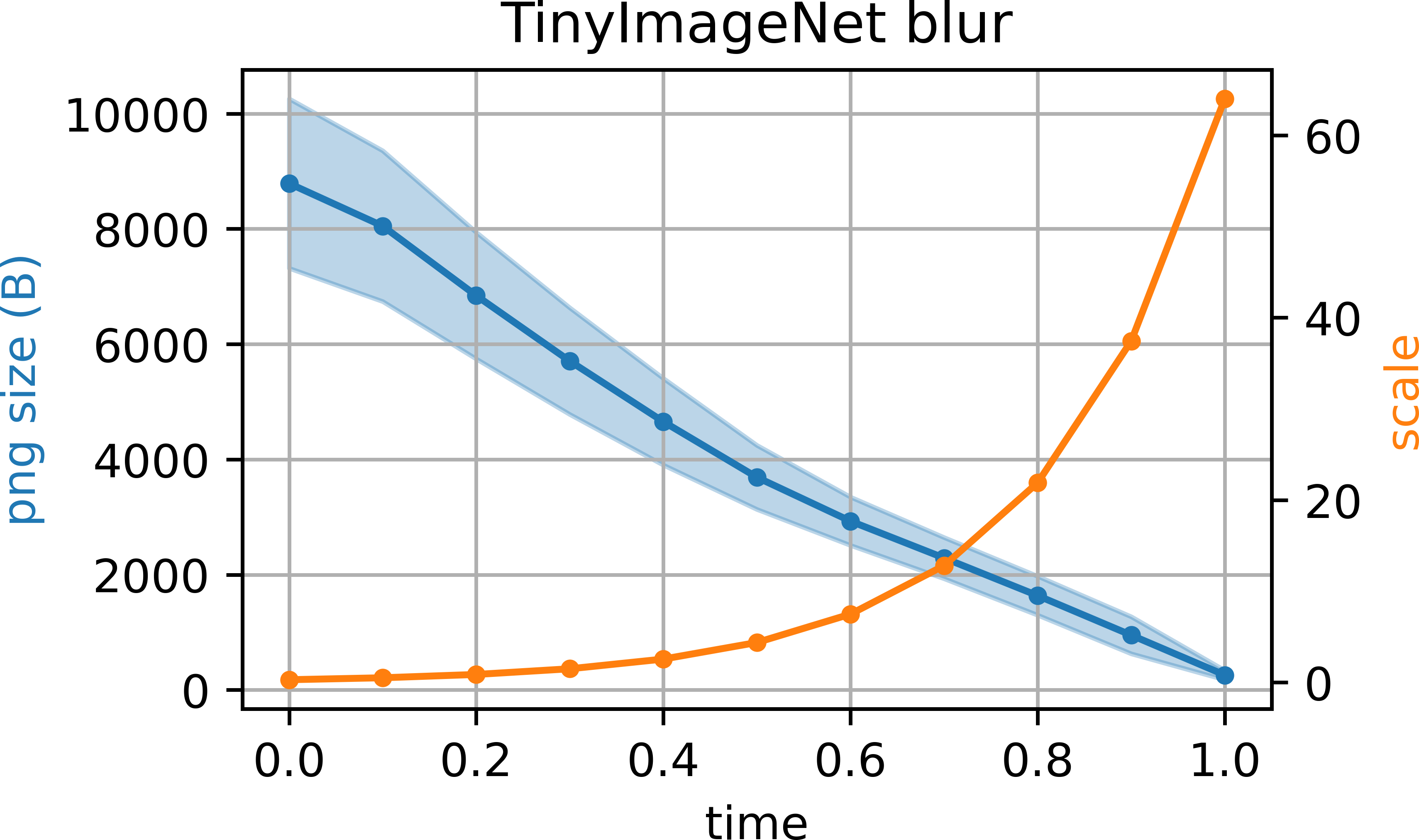}
    \caption{{Blur reduces information linearly}.} 
    \label{fig:blurschedule}

    \vspace{-3ex}
\end{figure}

\paragraph{Blurring.} We also follow standard practice in blurring for GDMs \citep{hoogeboom2022blurring}
\begin{align}
    \x_t^\mathrm{blur} &= \mathbf{V} \exp\big(-\tau(t) \boldsymbol{\Lambda}\big) \mathbf{V}^T \x,
\end{align}
where $\mathbf{V}^T$ is the discrete cosine transform (DCT), $\boldsymbol{\Lambda}$ are the squared frequencies $[\boldsymbol{\Lambda}]_{wh} = \pi^2( \frac{w^2}{W^2} + \frac{h^2}{H^2})$ of the pixel coordinates $(0,0) \le (w,h) < (W,H)$, and $\tau(t) \ge 0$ is a dissipation time (See Appendix A in \citet{rissanen2022generative} for detailed derivations). Blurring corresponds to `melting' the image by a heat equation $\Delta$ for $\tau$ time, which is equivalent to convolving the image with a heat kernel of scale $\sigma_B = \sqrt{2 \tau}$ \citep{rissanen2022generative}, and we fix dissipation time $\tau(t) = \sigma_B(t)^2/2$ to the scale.
We follow a logarithmic scale between $\sigma_\mathrm{min}=0.3$ and full image width $\sigma_\mathrm{max}=W$,
\begin{align}
    \sigma_B(t) &= \exp\Big( (1-t) \log \sigma_\mathrm{min} + t \log \sigma_\mathrm{max} \Big),
\end{align}
which has been found to reduce image information linearly \citep{rissanen2022generative}. We verify this in \cref{fig:blurschedule}, where we apply the Lempel-Ziv and Huffman compression of PNG codec\footnote{We use \texttt{torchvision.io.encode\_png}} to blurred TinyImageNet dataset to measure the information content post-blur. See \cref{fig:blur} for blur examples.

\subsection{Label Smoothing}
After discussing possible strategies for mollifying the inputs, we are ready to discuss strategies to define the likelihood function $p(\y_n | \x_n, \phi, \theta)$ for corrupted inputs.
We consider two common forms of label degradation,
\begin{align} \label{eq:label:smoothing}
    \y^{\mathrm{temp}}_t &= (1-\gamma_t) \y^{\mathrm{onehot}}, \qquad \gamma_t \in [0,1] \\
    \y^{\mathrm{LS}}_t &= (1-\gamma_t) \y^{\mathrm{onehot}} + \frac{\gamma_t}{C} \mathbf{1},
\end{align}
where $\gamma_t : [0,1] \mapsto [0,1]$ is a monotonic label decay. The labels $\y_t$ can be seen to follow a Dirac distribution $p(\y_t | \x_t, \y) = \delta(\y_t)$. 
In the $\y^{\mathrm{temp}}$ we simply reduce the true label down, while keeping the remaining labels at 0. This has been shown to be equivalent to a \emph{tempered} Categorical likelihood \citep{kapoor2022uncertainty}
\begin{align}
    p_\mathrm{temp}(\y | \f) &= \mathrm{Cat}( \y | \f)^{\gamma_t} = f_y^{\gamma_t}.
\end{align}
This can be seen as a `hot' tempering as $\gamma_t < 1$, where we flatten the likelihood surface. 

In contrast, in the $\y^{\mathrm{LS}}$ we perform \emph{label smoothing}, where we linearly mix one-hot and uniform label vectors \citep{szegedy2016rethinking,muller2019does}. Label smoothing can seem unintuitive, since we are adding value to all $C-1$ incorrect labels. Why should a noisy `frog' image have some `airplane' label?

\paragraph{Dirichlet interpretation.}
To shed light on this phenomenon, we draw a useful interpretation of the Categorical cross-entropy as a Dirichlet distribution
\begin{align}
    \log p(\y | \f) & \propto \log \mathrm{Dir}( \f | \1 + \y) \\
     &= \sum_c y_c \log f_c - \log B(\y),
\end{align}
for a normalized prediction probability vector $\f = (f_1, \ldots, f_C) := \f(\x)$. The Dirichlet distribution is a distribution over normalized probability vectors, which naturally is the probability vector $\f$, while the observations are proportional to the class \emph{concentrations} \citep{sensoy2018evidential}. A higher concentration favors higher probabilities for the respective classes. A standard cross-entropy likelihood for a one-hot label is proportional to Dirichlet with concentrations $(\ldots, 1, 2, 1, \ldots)$.

\begin{figure}[t]
    \centering
    \includegraphics[width=0.9\linewidth]{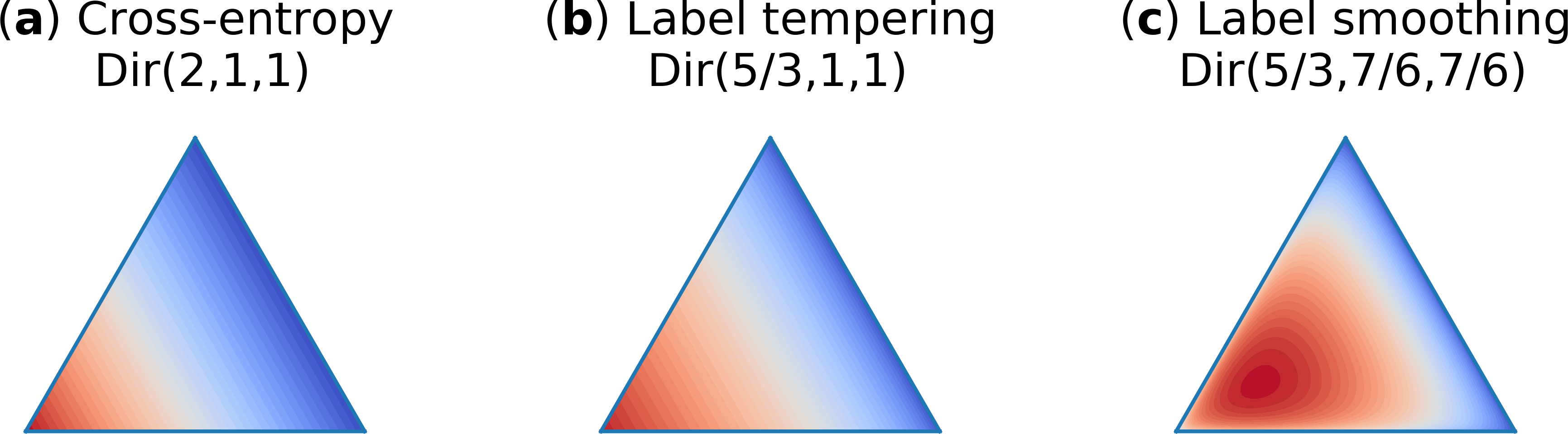}
    \caption{\textbf{Label smoothing is an intuitive way to degrade multi-class labels.} The Dirichlet $\mathrm{Dir}(\f | \1 + \y)$ visualizations show cross-entropy $\y^\mathrm{onehot}$ (a) and label tempering $\y^\mathrm{temp}$ (b) to retain prediction mode at $(1,0,0)$, while label smoothing $\y^\mathrm{LS}$ (c) prefers non-peaky predictions.}
    \label{fig:dirichlet}
    \vspace{-2ex}
\end{figure}

\cref{fig:dirichlet} compares the Dirichlet densities across clean $\y^{\mathrm{onehot}}$, tempered $\y^{\mathrm{temp}}$ and smoothed $\y^{\mathrm{LS}}$ labels. We notice that tempering the correct label down flattens the slope of the prediction density, but retains its mode at $\f^* = \y^\mathrm{onehot}$, which can lead to overfitting and poor calibration \citep{guo2017calibration} (See \cref{fig:dirichlet}c). In label smoothing, while increasing the density associated with the wrong labels in $\y^{\mathrm{LS}}$ can seem counter-intuitive, it can be shown to favor reduced prediction confidences for corrupted images with a mode $\f^* = \y^{\mathrm{LS}}$ (See \cref{fig:dirichlet}b). The wrong labels are induced to have equal logits.

\paragraph{Label noising schedule.}

Finally, we need to decide how quickly the labels are smoothed given an input $\x_t$. For small temperatures (e.g., $t < 0.1$) we expect close to no effect on the label, while for high temperature (e.g., $t > 0.9$) we expect the label to almost disappear. For noise-based input mollification, we propose a label decay dependent on the signal-to-noise ratio $\mathrm{SNR}(t) = \alpha_t^2/\sigma_t^2$ as follows \citep{kingma2021variational}:
\begin{align}
    \gamma_t^\mathrm{noise} &= \left(\frac{1}{1 + \alpha_t^2/\sigma_t^2 }\right)^k.
\end{align}
That is, when the image has equal amounts of signal and noise with $\mathrm{SNR}(t) = 1$, we assume that half of the label remains; $k$ is a slope hyperparameter, which we estimate empirically.

For blurring, the SNR is undefined as blurry images contain no added noise. Instead, we propose to apply label smoothing by the number of bits of information in the blurry images, measured through an image compression algorithm. 
\cref{fig:blurschedule} offers interesting insights on the reduction of information associated with the blurring intensity. 
Based on this analysis, we define a linear label smoothing approximation
\begin{align}
    \gamma_t^\mathrm{blur} &= \left( \frac{\operatorname{size}( \x_t\texttt{.png} ) }{\operatorname{size}( \x_0\texttt{.png} )} \right)^k \approx t^k,
\end{align}
where we apply label smoothing by the bit-ratio, and where $k$ is a hyperparameter determining either a faster ($k<1$) or slower ($k>1$) label decay.

\newcolumntype{a}{>{\columncolor{cyan!5!white}}c}
\begin{table*}[tb!]
\begin{center}
\caption{Results on preactResNet-50 ($23.7$M params.). FCR: horizontal flips, crops and rotations. A tick mark for ``Moll.'' indicates the use of the proposed data mollification.}
\label{tab:benchmark2}
\vspace{-1.75ex}
\resizebox{0.97\linewidth}{!}{
\setlength{\tabcolsep}{4pt}
\begin{tabular}{lc aaccaa ccaacc aaccaa}
& & \multicolumn{6}{c}{CIFAR-10} & \multicolumn{6}{c}{CIFAR-100} & \multicolumn{6}{c}{TinyImageNet} \\
\cmidrule(lr){3-8} \cmidrule(lr){9-14} \cmidrule(lr){15-20}
& & \multicolumn{2}{c}{Error $(\downarrow)$} & \multicolumn{2}{c}{NLL $(\downarrow)$} & \multicolumn{2}{c}{ECE $(\downarrow)$} & \multicolumn{2}{c}{Error} & \multicolumn{2}{c}{NLL} & \multicolumn{2}{c}{ECE} & \multicolumn{2}{c}{Error} & \multicolumn{2}{c}{NLL} & \multicolumn{2}{c}{ECE} \\
\cmidrule(lr){3-4} \cmidrule(lr){5-6} \cmidrule(lr){7-8} \cmidrule(lr){9-10} \cmidrule(lr){11-12} \cmidrule(lr){13-14} \cmidrule(lr){15-16} \cmidrule(lr){17-18} \cmidrule(lr){19-20}
\rowcolor{white} Augmentation & Moll. & clean & corr & clean & corr & clean & corr & clean & corr & clean & corr & clean & corr & clean & corr & clean & corr & clean & corr \\
\toprule
(none) &\xmark & 11.7 & 29.2 & 0.48 & 1.35 & 0.09 & 0.22 & 39.5 & 59.9 & 1.67 & 2.85 & 0.15 & 0.23 & 53.7 & 83.1 & 2.51 & 4.22 & 0.15 & 0.12 \\ 
FCR &\xmark & 5.0 & 21.9 & 0.22 & 1.26 & 0.04 & 0.18 & 23.2 & 47.4 & 0.97 & 2.30 & 0.12 & 0.21 & 33.5 & 75.7 & 1.51 & 4.10 & 0.12 & 0.21 \\ 
FCR + RandAug &\xmark & 4.4 & 15.5 & 0.18 & 0.75 & 0.04 & 0.12 & 21.5 & 41.0 & 0.87 & 2.09 & 0.11 & 0.19 & 32.1 & 70.7 & \textbf{1.43} & 3.89 & 0.12 & 0.22 \\ 
FCR + AutoAug &\xmark & 4.5 & 13.2 & 0.19 & 0.60 & 0.04 & 0.10 & 23.3 & 39.2 & 0.93 & 1.89 & 0.11 & 0.18 & 33.6 & 69.0 & 1.49 & 3.74 & 0.12 & 0.21 \\ 
FCR + AugMix &\xmark & 4.5 & 10.6 & 0.18 & 0.43 & 0.04 & 0.08 & 22.8 & 35.3 & 0.94 & 1.59 & 0.12 & 0.16 & 34.0 & 63.5 & 1.51 & 3.38 & 0.12 & 0.20 \\ 
FCR + TrivAug & \xmark & \textbf{3.4} & 12.1 & \textbf{0.12} & 0.49 & \textbf{0.03} & 0.09 & 20.5 & 36.9 & \textbf{0.79} & 1.69 & \textbf{0.10} & 0.16 & 33.8 & 70.4 & 1.58 & 4.71 & 0.13 & 0.30 \\ 
FCR + MixUp &\xmark & 4.1 & 21.9 & 0.32 & 0.84 & 0.19 & 0.22 & 21.9 & 46.1 & 1.01 & 2.11 & 0.18 & 0.19 & 39.3 & 76.9 & 1.86 & 3.90 & 0.16 & 0.16 \\ 
FCR + CutMix &\xmark & 4.2 & 28.4 & 0.20 & 1.25 & 0.09 & 0.18 & 22.6 & 54.2 & 1.12 & 2.83 & 0.18 & 0.23 & 36.1 & 79.8 & 1.65 & 4.51 & 0.12 & 0.26 \\ 
\midrule 
(none) & \cmark & 15.8 & 21.4 & 0.68 & 0.87 & 0.12 & 0.14 & 49.3 & 54.5 & 2.28 & 2.53 & 0.14 & 0.14 & 50.3 & 67.4 & 2.44 & 3.41 & 0.15 & \textbf{0.15} \\ 
FCR & \cmark & 5.8 & 10.2 & 0.25 & 0.43 & 0.04 & 0.08 & 27.1 & 34.7 & 1.17 & 1.57 & 0.11 & 0.13 & 35.0 & 60.4 & 1.63 & 3.07 & 0.12 & \textbf{0.15} \\ 
FCR + RandAug & \cmark & 4.4 & 8.3 & 0.18 & 0.34 & 0.04 & \textbf{0.07} & 23.0 & 30.5 & 0.96 & 1.32 & \textbf{0.10} & \textbf{0.12} & 33.2 & 57.4 & 1.49 & 2.87 & \textbf{0.11} & \textbf{0.15} \\ 
FCR + AutoAug & \cmark & 4.4 & 8.2 & 0.17 & \textbf{0.32} & 0.04 & 0.08 & 23.4 & 30.6 & 0.95 & 1.31 & \textbf{0.10} & 0.13 & 33.8 & 57.8 & 1.52 & 2.88 & \textbf{0.11} & 0.16 \\ 
FCR + AugMix & \cmark & 5.3 & 8.6 & 0.21 & 0.35 & 0.04 & 0.08 & 25.9 & 32.5 & 1.08 & 1.42 & 0.11 & 0.13 & 36.2 & \textbf{56.9} & 1.65 & \textbf{2.84} & 0.12 & 0.15 \\ 
FCR + TrivAug & \cmark & 4.0 & \textbf{8.0} & 0.15 & \textbf{0.32} & 0.04 & \textbf{0.07} & \textbf{20.4} & \textbf{28.5} & 0.81 & \textbf{1.19} & \textbf{0.10} & \textbf{0.12} & 32.2 & 59.2 & \textbf{1.43} & 3.05 & \textbf{0.11} & 0.17 \\ 
FCR + MixUp & \cmark & 4.7 & 9.4 & 0.35 & 0.51 & 0.21 & 0.22 & 22.5 & 31.8 & 1.08 & 1.47 & 0.18 & 0.18 & 34.3 & 61.7 & 1.81 & 3.13 & 0.23 & 0.18 \\ 
FCR + CutMix & \cmark & 4.1 & 10.9 & 0.25 & 0.47 & 0.14 & 0.16 & 22.6 & 34.8 & 1.11 & 1.65 & 0.19 & 0.18 & \textbf{31.0} & 64.8 & 1.62 & 3.37 & 0.19 & 0.17 \\ 
\bottomrule 

\end{tabular}
}
\end{center}
\vspace{-2.5ex}
\end{table*}

\subsection{Cross-entropy likelihood}

Our analysis points towards label smoothing as a suitable candidate to define a perturbation of the labels associated with mollified inputs, and we propose a cross-entropy-based likelihood
\begin{align} \label{eq:crossentropy_likelihood}
    \log p(\y | \x, \phi, \theta) &= \log p(\y_t^{\mathrm{LS}} | \x_t ; \theta)\\
    &= \sum_{c=1}^C \y_{t,c}^{\mathrm{LS}} \log f(\x_t)_c, %
\end{align}
where $c$ are the class indices. 
Using an un-normalized cross-entropy is standard practice in label smoothing \citep{muller2019does}. We note that defining the likelihood with a Dirichlet has a different meaning, since the Dirichlet distribution is normalized such that $\int \mathrm{Dir}(\f | \y) d\f = 1$, while a proper likelihood is normalized w.r.t. the data domain $\y$ (see \citet{kapoor2022uncertainty,wenzel20posterior} for a discussion).

\begin{figure*}[htb]
    \centering
    \includegraphics[width=0.91\textwidth]{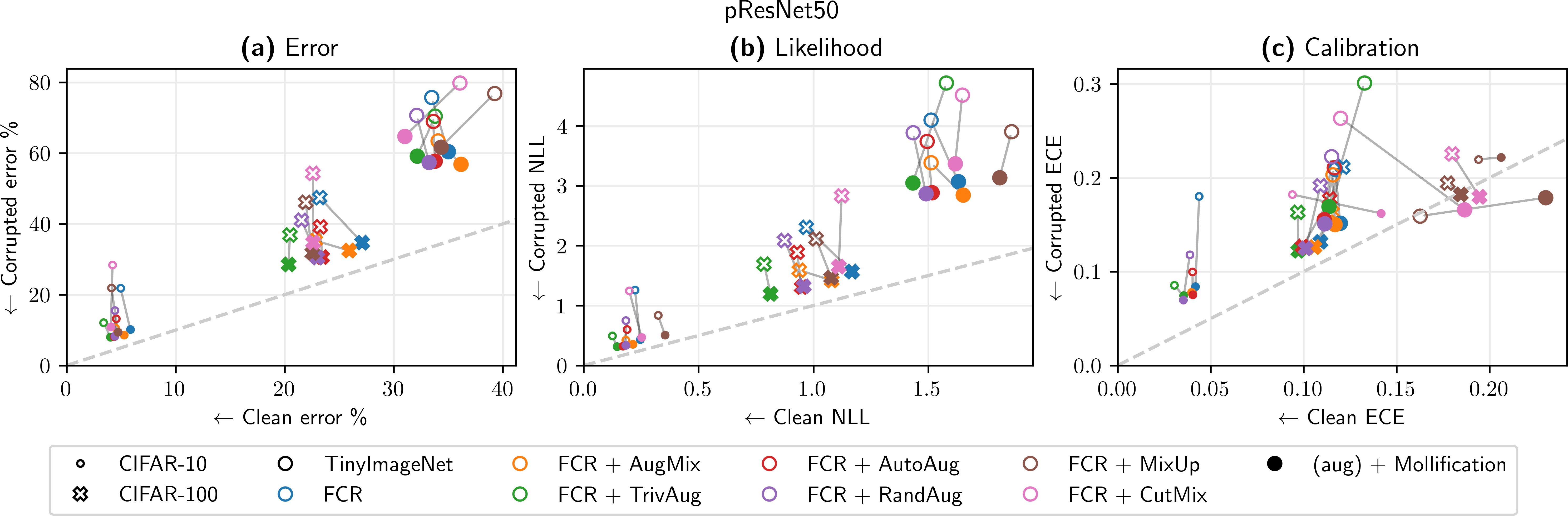}
    \caption{Adding mollification (\protect\tikz\protect\draw[blue,fill=blue] (0,0) circle (.5ex); $\cdots$\protect\tikz\protect\draw[pink,fill=pink] (0,0) circle (.5ex);) to augmentations (\protect\tikz\protect\draw[blue] (0,0) circle (.5ex); $\cdots$\protect\tikz\protect\draw[pink] (0,0) circle (.5ex);) improves corrupted accuracy (\textbf{a}), likelihood (\textbf{b}) and calibration (\textbf{c}) over CIFAR-10, CIFAR-100 and TinyImageNet.} 
    \label{fig:errcorr}
    \vspace{-3ex}
\end{figure*}

\begin{figure*}[htb]
     \centering
     \subfloat[][Mollification type]{\includegraphics[width=0.28\textwidth]{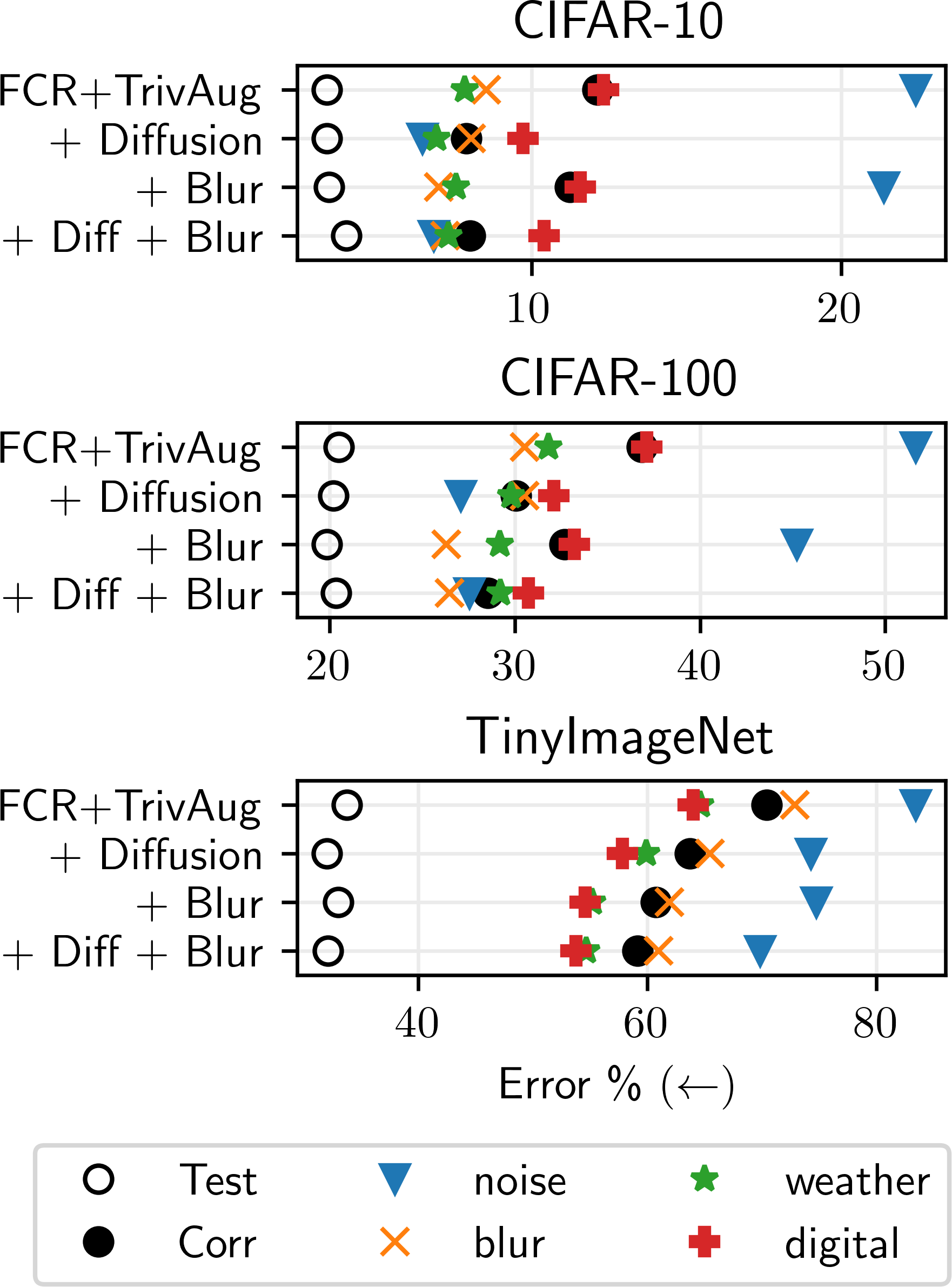}} 
     \hspace{0.5ex}
     \subfloat[][Amount of mollification]{\includegraphics[width=0.28\textwidth]{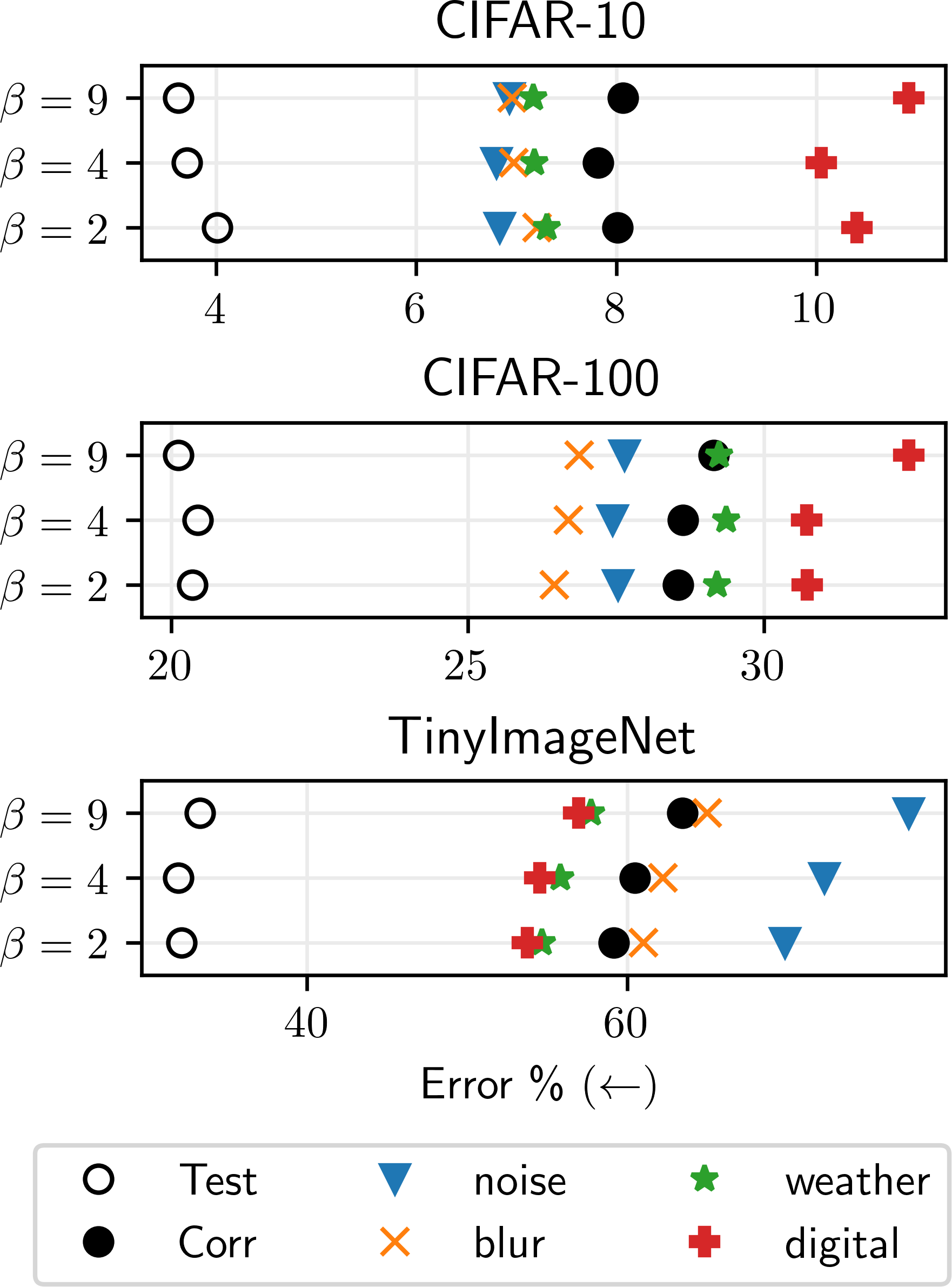}}
     \hspace{0.5ex}
     \subfloat[][Label smoothing schedule]{\includegraphics[width=0.28\textwidth]{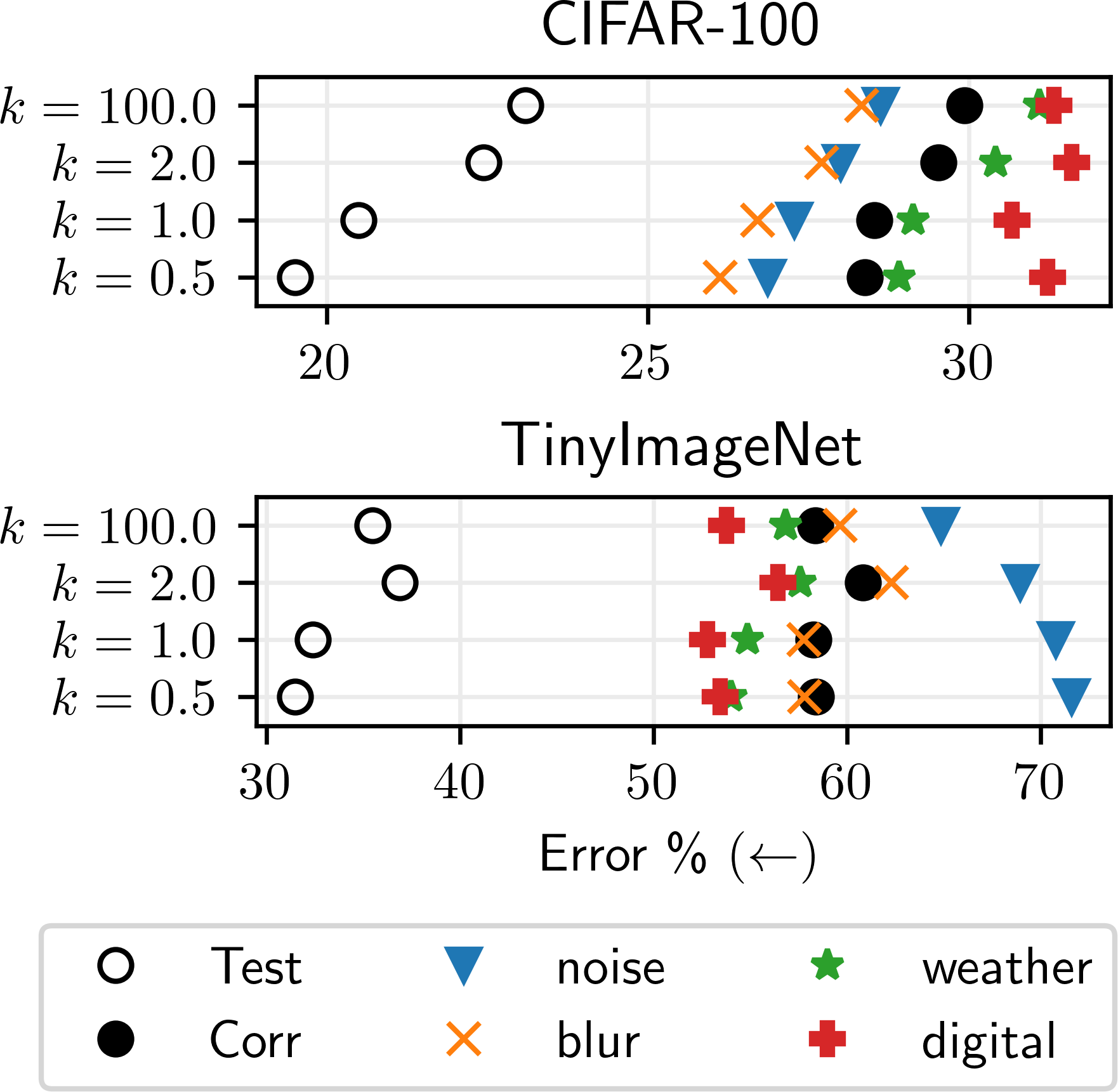}}
    \vspace{-1ex}
    \caption{{Combining substantial noising and blurring with greedy label smoothing yields good results}.} %
    \label{fig:ablation}
\end{figure*}

\section{EXPERIMENTS}
\label{sec:exp}

\begin{table*}[h]
\begin{center}
\caption{Breakdown of errors over individual corruption types for a pResNet-50 network.}
\label{tab:breakdown}
\vspace{-2ex}
\resizebox{0.9\linewidth}{!}{
\setlength{\tabcolsep}{3pt}
\begin{tabular}{ l l c ccc cccc cccc cccc c } 
\toprule 
 & & & \multicolumn{3}{c}{noise} & \multicolumn{4}{c}{blur} & \multicolumn{4}{c}{weather} & \multicolumn{4}{c}{digital} &  \\ 
\cmidrule(lr){4-6} \cmidrule(lr){7-10} \cmidrule(lr){11-14} \cmidrule(lr){15-18} 
& & Clean  & shot & impulse & gauss & motion & zoom & defocus & glass & fog & frost & snow & bright & jpg & pixel & elastic & contrast & mean \\ 
 \midrule 
\multirow{4}{*}{\rotatebox{90}{CIFAR10}} & FCR+TrivAug & \textbf{3.4} & 23 & 13 & 31 & 10.0 & 6.2 & 5.5 & 12 & \textbf{6.3} & 12 & 9.3 & \textbf{3.9} & 16 & 21 & 7.4 & \textbf{4.4} & 12 \\ 
 & + Diffusion & \textbf{3.4} & \textbf{5.7} & \textbf{7.3} & \textbf{6.4} & 9.4 & 6.6 & 5.8 & \textbf{10} & 6.8 & \textbf{8.4} & \textbf{8.6} & \textbf{4.0} & \textbf{12} & \textbf{15} & 7.2 & 4.9 & \textbf{7.6} \\ 
 & + Blur & 3.5 & 20 & 17 & 27 & \textbf{8.1} & \textbf{4.8} & \textbf{4.1} & 11 & \textbf{6.4} & 11 & 9.2 & 4.0 & 17 & 18 & \textbf{6.5} & 4.7 & 11 \\ 
 & + Diff+Blur & 4.0 & 6.2 & 7.5 & 6.8 & 8.8 & 5.0 & 4.6 & \textbf{10} & 7.1 & \textbf{8.5} & 9.1 & 4.5 & 14 & 16 & 6.7 & 5.3 & 7.8 \\ 
 \midrule 
 \multirow{4}{*}{\rotatebox{90}{CIFAR100}} & FCR+TrivAug & \textbf{20} & 54 & 39 & 63 & 32 & 28 & 25 & 38 & 30 & 41 & 33 & 23 & 47 & 47 & 30 & \textbf{25} & 36 \\ 
 & + Diffusion & \textbf{20} & \textbf{25} & \textbf{30} & \textbf{26} & 32 & 29 & 27 & 34 & 31 & 34 & 32 & 23 & \textbf{37} & 36 & 29 & 26 & 29 \\ 
 & + Blur & \textbf{20} & 45 & 40 & 51 & \textbf{28} & \textbf{23} & \textbf{21} & 33 & \textbf{29} & 36 & \textbf{31} & \textbf{22} & 46 & 36 & \textbf{26} & \textbf{25} & 32 \\ 
 & + Diff+Blur & \textbf{20} & \textbf{25} & 31 & \textbf{26} & \textbf{28} & 23 & 22 & \textbf{32} & 30 & \textbf{33} & \textbf{31} & 23 & 39 & \textbf{32} & 27 & 25 & \textbf{28} \\ 
 \midrule 
 \multirow{4}{*}{\rotatebox{90}{TIN}} & FCR+TrivAug & 34 & 81 & 85 & 85 & 68 & 71 & 76 & 76 & 65 & 65 & 69 & 60 & 59 & 61 & 61 & 76 & 68 \\ 
 & + Diffusion & \textbf{32} & 70 & 78 & 75 & 61 & 62 & 68 & 70 & 61 & 60 & 64 & 55 & 52 & 55 & 53 & 71 & 62 \\ 
 & + Blur & 33 & 71 & 77 & 77 & \textbf{57} & 58 & 64 & \textbf{68} & \textbf{56} & 56 & \textbf{60} & \textbf{50} & 51 & \textbf{49} & \textbf{49} & 69 & 59 \\ 
 & + Diff+Blur & \textbf{32} & \textbf{65} & \textbf{73} & \textbf{71} & \textbf{57} & \textbf{57} & \textbf{63} & \textbf{67} & \textbf{56} & \textbf{54} & \textbf{59} & \textbf{50} & \textbf{50} & \textbf{49} & \textbf{49} & \textbf{68} & \textbf{57} \\ 
 \bottomrule 
\end{tabular} 
}
\end{center}
\end{table*}

\paragraph{Datasets.}
We analyze the effect of input mollification and label smoothing, which we collectively refer to as {\em data mollification}, on the CIFAR-10, CIFAR-100  \citep{krizhevsky2009cifar} and TinyImageNet-200 \citep{Le2015TinyIV} datasets. The CIFAR datasets contain $50$K training images and $10$K validation images at $32\times32$ resolution with 10 or 100 classes. The TinyImageNet dataset contains 100K training images with $10$K validation images at $64\times64$ resolution over $200$ classes. By default we apply training-time augmentations of horizontal flips, random cropping (padding of $4$ pixels) and random rotations (up to $15$ degrees). We compare six augmentations of AugMix, RandAug, AutoAug, TrivAug, MixUp and CutMix with and without data mollification. We implemented these on PyTorch Lightning, and ran experiments on individual NVIDIA V100 GPUs. %
The code is available at \url{github.com/markusheinonen/supervised-mollification}.

\paragraph{Training.}
We use a preact-ResNet-50 \citep{he2016identity} with $23.7$ million parameters
\footnote{Experiments on other network architectures yielded similar results and are available in \cref{sec:arch}.}. We train with SGD for 300 epochs using an initial learning rate of $0.01$, which follows a Cosine annealing schedule \citep{loshchilov2016sgdr}. We use mini-batches of size 128, and the cross-entropy-based likelihood. Mollification incurs a negligible running time increase, similar to other augmentations. 
We find that simple augmentations of flips (F), crops (C) and rotations (R) are always necessary to achieve satisfying performance, even with mollification.

\begin{wrapfigure}[12]{r}{0.45\columnwidth}
    \centering
    \includegraphics[width=0.45\columnwidth]{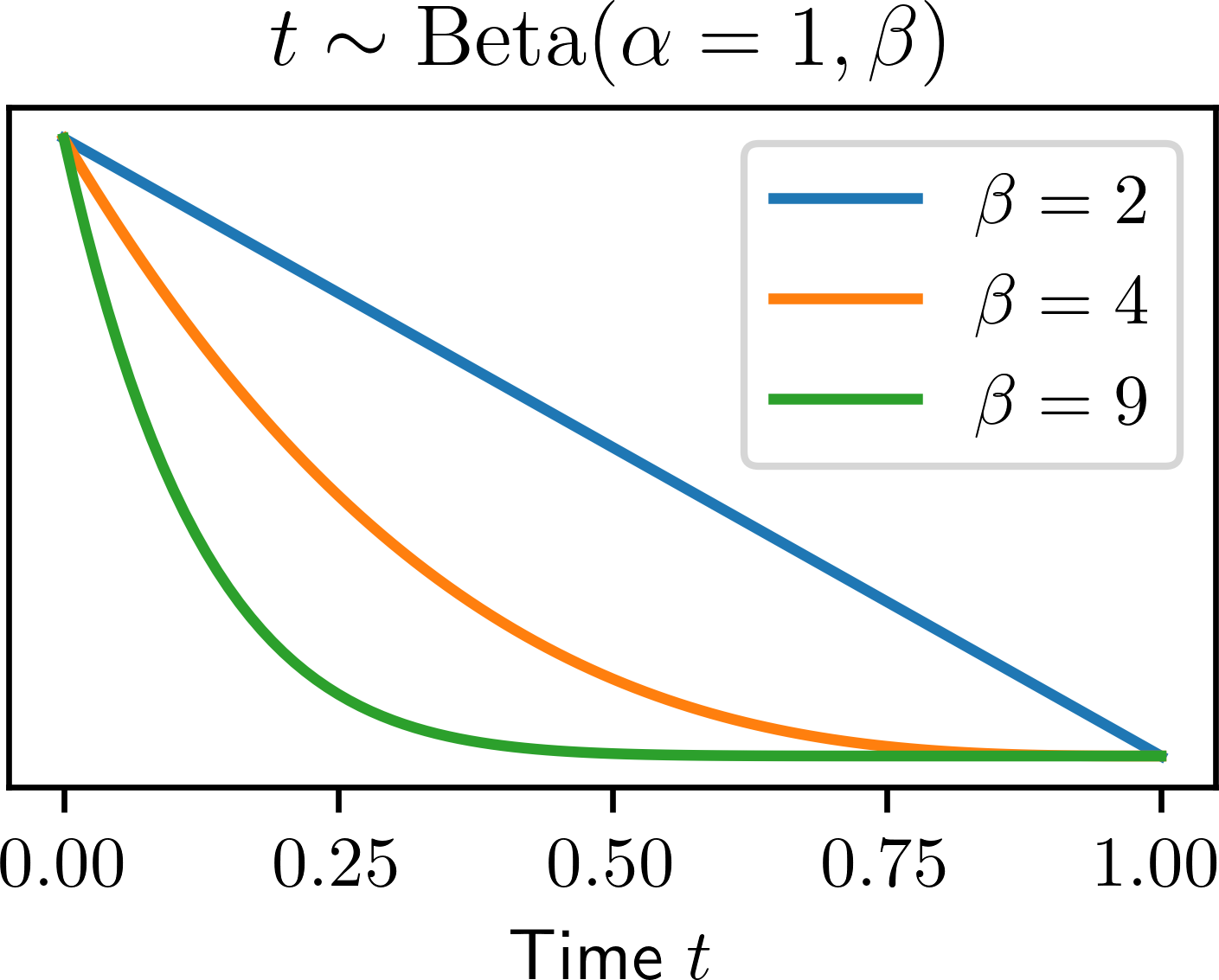}
    \caption{Temperature distributions.}
    \label{fig:beta}
\end{wrapfigure}
\paragraph{Mini-batching.} During training we randomly pick the transformation (noising, burring, or no mollification) for each image $\x_n$ in a mini-batch. 
For the noisy and blurry images we further sample separate temperatures $t_n \sim \mathrm{Beta}(\alpha,\beta)$ from a Beta distribution per image (see \cref{fig:beta} for an illustration). We choose by default $\alpha=1$ and $\beta=2$, which results in  average temperature $\E[t] = 1/3$.

\paragraph{Metrics}
We report error, negative log-likelihood $-\log p(\y|\f(\x))$ (NLL), and expected calibration error (ECE) \citep{guo2017calibration} on the validation set as well as on corrupted validation images from CIFAR-10-C, CIFAR-100-C and TinyImageNet-C, which contain $75$ corruptions of each validation image using $15$ types of corruptions at $5$ corruption magnitude levels \citep{hendrycks2019benchmarking}. The corruptions divide in four categories of noises (3/15), blurs (4/15), weather effects (3/15), and colorization or compression artifacts (5/15). %

\subsection{Benchmark results}

\begin{figure*}[h]
    \centering
    \includegraphics[width=0.85\textwidth]{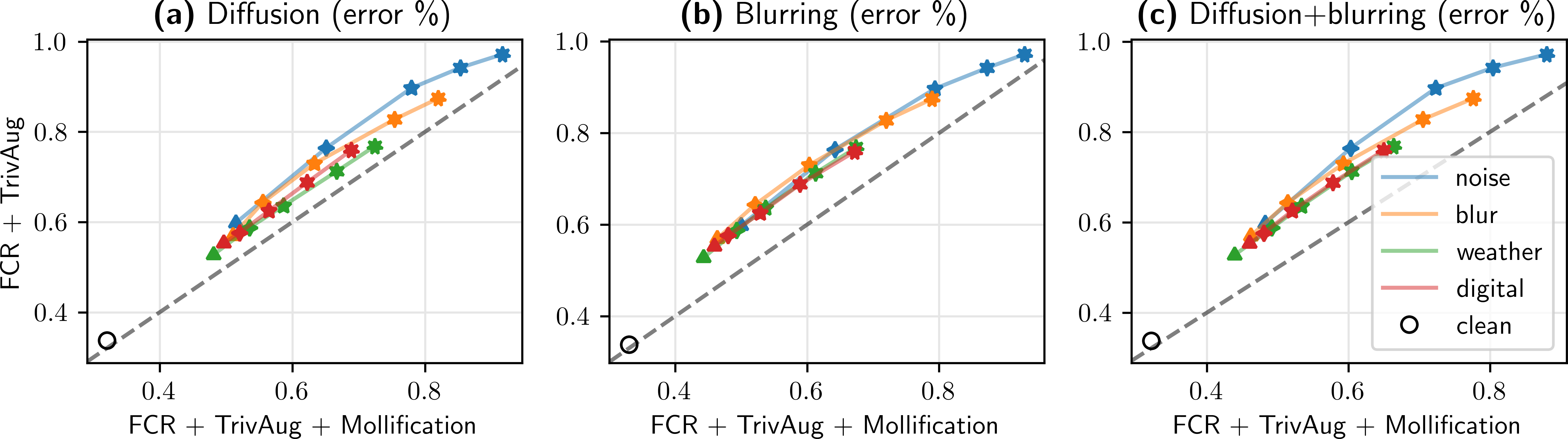}
    \caption{{Blurring-based input mollification improves error against all corruption types evenly, while incorporating noise-based mollification adds slight benefits}. Polygon size indicates corruption severity from triangles (1) to heptagons (5); experiment on TinyImageNet.}
    \label{fig:corrtypes}
    \vspace{-2ex}
\end{figure*}

\begin{figure*}[h]
    \centering
    \includegraphics[width=0.9\textwidth]{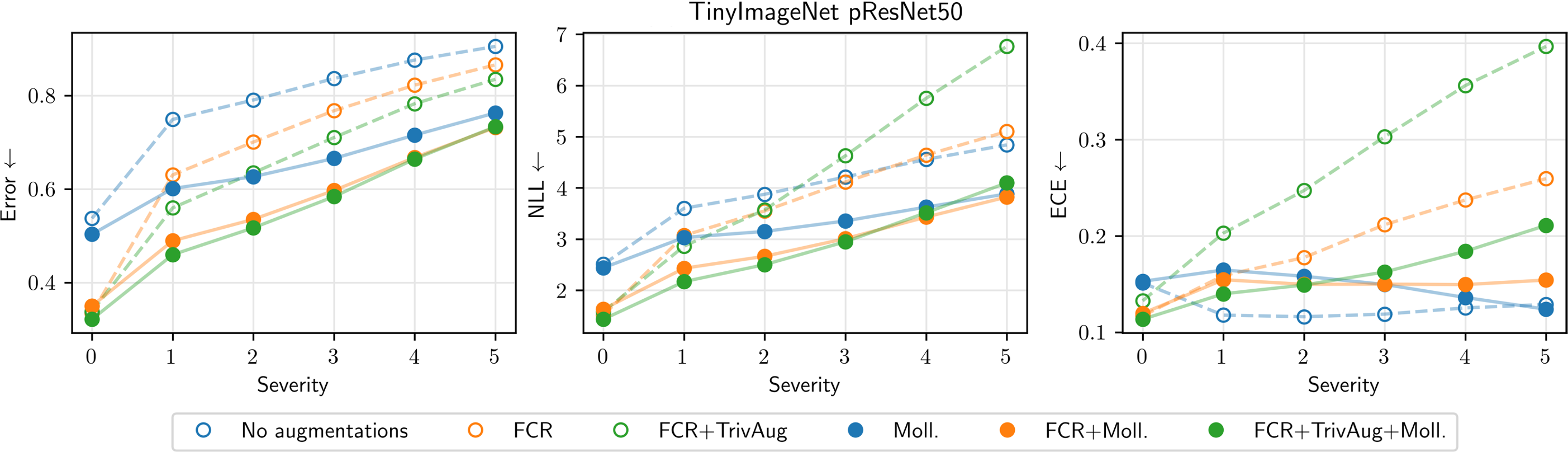}
    \caption{Data mollification improves augmented models over the corruption severity.} 
    \label{fig:severity}
    \vspace{-2ex}
\end{figure*}

We report benchmark results in \cref{tab:benchmark2}. We first observe that among the classic augmentation techniques TrivAug combined with FCR is superior with errors of 3.4\%, 20.5\% and 33.8\% in CIFAR-10, CIFAR-100 and TinyImageNet, respectively. Notably, we struggled to achieve good performance with MixUp and CutMix. 

When we include mollification, we again observe that TrivAug+Mollification yields highest performance, while mollification improves corrupted image errors on all 6 augmentations on all datasets. On CIFAR-10 we improve corrupted error $12.1\% \rightarrow 8.0\%$, on CIFAR-100 $36.9\% \rightarrow 28.5\%$ and on TinyImageNet $70\% \rightarrow 59\%$. Mollification also improves the clean performance on the larger TinyImageNet ($33.8\% \rightarrow 32.2\%$), has little effect on CIFAR-100, and slightly decreases on CIFAR-10.
We also observe minor calibration improvements from mollification on clean images, while there are some 20\% to 40\% improvements on calibration on corrupted validation sets. 

\cref{fig:errcorr} shows the relationship between performance on the clean and corrupted validation sets, and it indicates that mollification consistently narrows the gap between clean and corrupted metrics on all datasets. 
{
Although our method may in some cases slightly degrade the accuracy on clean data, it consistently and greatly enhances robustness against corruptions and improves uncertainty estimates in most cases, as evidenced by improved performance in terms of NLL and ECE.
}

{
We observe that blurring works better on higher-resolution data like TinyImageNet, but it is less effective than noisy augmentation for lower-resolution data like CIFAR-10. 
We attribute this to CIFAR-10 already being rather down-scaled so that blurring removes label information quickly. 
In contrast, TinyImageNet images can handle higher blur intensities while retaining label information.
A potential future work is to explore resolution-based blur and label smoothing schedules.
}

\subsection{Ablation: How to choose mollification and its intensity?}

Next, we study the hyperparameters governing mollification. \cref{fig:ablation}(a) shows that on smaller CIFAR images noise diffusion improves more than blurring, while on larger TinyImageNet blurring is more advantageous (see \cref{fig:corrtypes}). Combining both modes during training is superior in all cases. Notably, training under blur also improves robustness to test noising on TinyImageNet. We also observe that mollification gives highest performance improvements to `weather' and `digital' test corruptions, despite not encoding for them during training time. \cref{fig:ablation}(b) shows that larger mollification amounts (with $\beta=2$) gives the best performance, while panel \cref{fig:ablation}(b) shows the benefits in choosing more aggressive label decay with $k \le 1$.

\subsection{Effect of corruption severity}

We visualize the performance w.r.t. corruption severity in \cref{fig:severity} for the FCR + TrivAug base augmentation, for which higher values result in higher errors. 
The mollified versions are in almost all cases better than the corresponding non-mollified ones on all corruption intensities, while giving similar performance on clean images (zero severity).

\subsection{Which corruptions do we improve?}

The corruptions in the corrupted datasets fall into four categories. \cref{fig:ablation} shows a general trend of the network being most robust against `digital' and `weather' corruptions, while `blur' corruptions are moderately difficult, and `noise' corruptions are most difficult to predict correctly.
{
In \cref{sec:spectral_analysis}, we provide a spectral analysis explanation for this phenomenon.
}
\cref{tab:breakdown} shows the performance w.r.t. the 15 individual corruptions over the three datasets. Mollification improves consistently: in TinyImageNet all individual corruption types improve between 9 and 16 percentage points from TrivAug, which is the best performing conventional augmentation method.

\subsection{Results on ImageNet}

\begin{table*}[h]
\begin{center}
\caption{Breakdown of errors over individual corruption types for a ResNet-50 network on ImageNet. The numbers in the table are the differences between the errors obtained with and without mollification, so negative numbers indicate improvements of mollification with respect to baselines without mollification.}
\label{tab:imagenetC:breakdown}
\resizebox{1.0\linewidth}{!}{
\setlength{\tabcolsep}{3pt}
\begin{tabular}{ l l c ccc cccc cccc cccc c } 
\toprule 
 & & & \multicolumn{3}{c}{noise} & \multicolumn{4}{c}{blur} & \multicolumn{4}{c}{weather} & \multicolumn{4}{c}{digital} &  \\ 
\cmidrule(lr){4-6} \cmidrule(lr){7-10} \cmidrule(lr){11-14} \cmidrule(lr){15-18} 
& Moll. & Clean  & shot & impulse & gauss & motion & zoom & defocus & glass & fog & frost & snow & bright & jpg & pixel & elastic & contrast & mean \\ 
 \midrule 
\multirow{8}{*}{\rotatebox{90}{ImageNet}} 
& (none)         &  0.4 & -13.7 & -15.8 & -14.8 & -1.5 & -1.0 & -13.4 & -1.8 & -4.4 & -0.1 & 0.5 & 0.5 & 1.1 & -2.3 & -1.2 & -2.5 & -4.7       \\
& FCR            &  0.5 & -11.7 & -13.8 & -12.5 & -0.6 & -0.4 & -13.5 & -1.6 & -3.3 & 0.2 & 0.1 & 0.7 & 1.1 & -1.7 & -1.6 & -1.6 & -4.0        \\
& FCR + RandAug  &  0.5 & -12.1 & -13.2 & -13.5 & -2.1 & -1.3 & -13.4 & -3.1 & -4.7 & -0.5 & -1.1 & -0.6 & 1.6 & -3.0 & -2.8 & -1.6 & -4.8     \\
& FCR + AutoAug  &  0.5 & -11.0 & -12.7 & -12.4 & -2.5 & -2.0 & -13.4 & -2.4 & -3.7 & -1.9 & -0.3 & -0.8 & 1.5 & -1.3 & -2.1 & -1.9 & -4.5     \\
& FCR + AugMix   &  1.2 & -9.8 & -13.4 & -11.5 & 0.2 & -0.3 & -11.0 & -2.0 & -2.6 & -0.5 & -0.6 & 0.2 & 0.6 & -2.0 & -1.6 & -1.2 & -3.7        \\ 
& FCR + TrivAug  &  0.2 & -8.9 & -11.8 & -10.7 & -1.4 & 0.0 & -12.5 & -1.5 & -2.1 & 0.1 & -0.4 & -0.3 & 1.4 & -1.2 & -1.2 & -0.9 & -3.4        \\
& FCR + MixUp    &  0.1 & -13.7 & -14.9 & -13.9 & -2.6 & -2.5 & -15.8 & -2.7 & -4.0 & -0.7 & -0.8 & -0.8 & 0.6 & -1.8 & -1.9 & -4.6 & -5.3     \\
& FCR + CutMix   &  0.5 & -12.4 & -13.6 & -13.3 & -1.4 & -1.2 & -12.2 & -1.6 & -3.4 & -0.3 & -0.9 & 0.0 & 1.1 & -2.3 & -1.0 & -2.1 & -4.3      \\
\bottomrule 
\end{tabular} 
}
\end{center}
\end{table*}

We evaluate the proposed mollification method on the ImageNet dataset. 
Due to the large size of this dataset, we decide to perform a variation on the experimental setup compared to the other datasets. 
In particular, we consider a pretrained ResNet50 architecture and we fine-tune it for 10 epochs instead of training the model from scratch.

\newcolumntype{a}{>{\columncolor{cyan!5!white}}c}
\begin{table}[h]
\begin{center}
\caption{Results on ImageNet with a ResNet-50 architecture. FCR: horizontal flips, crops and rotations. A tick mark for ``Moll.'' indicates the use of the proposed data mollification.}
\label{tab:imagenet_clean}
\resizebox{1.0\linewidth}{!}{
\setlength{\tabcolsep}{4pt}
\begin{tabular}{lc aaccaa}
& & \multicolumn{6}{c}{ImageNet} \\ 
\cmidrule(lr){3-8} 
& & \multicolumn{2}{c}{Error $(\downarrow)$} & \multicolumn{2}{c}{NLL $(\downarrow)$} & \multicolumn{2}{c}{ECE $(\downarrow)$} \\
\cmidrule(lr){3-4} \cmidrule(lr){5-6} \cmidrule(lr){7-8}
\rowcolor{white} Augmentation & Moll. & clean & corr & clean & corr & clean & corr \\
\toprule
(none) &\xmark         & 24.4 & 61.4 & 0.95 & 3.27 & 0.11 & 0.14 \\ 
FCR &\xmark            & 24.1 & 61.2 & 0.95 & 3.23 & 0.10 & 0.14 \\ 
FCR + RandAug &\xmark  & 24.9 & 60.2 & 0.97 & 3.14 & 0.10 & 0.14 \\ 
FCR + AutoAug &\xmark  & 25.2 & 60.2 & 0.99 & 3.11 & 0.10 & 0.13 \\ 
FCR + AugMix &\xmark   & 24.1 & 59.9 & 0.96 & 3.10 & 0.10 & 0.13 \\ 
FCR + TrivAug & \xmark & 24.9 & 58.2 & 0.96 & 2.96 & 0.10 & 0.13 \\ 
FCR + MixUp &\xmark    & 26.1 & 60.8 & 1.24 & 3.27 & 0.19 & 0.15 \\ 
FCR + CutMix &\xmark   & 25.0 & 62.3 & 1.14 & 3.35 & 0.17 & 0.14 \\ 
\midrule 
\bottomrule 
(none) &\cmark         & 24.8 & 56.7 & 1.01 & 3.08 & 0.11 & 0.16 \\ 
FCR &\cmark            & 24.6 & 57.1 & 1.00 & 3.10 & 0.11 & 0.16 \\ 
FCR + RandAug &\cmark  & 25.3 & 55.5 & 1.01 & 2.93 & 0.10 & 0.16 \\ 
FCR + AutoAug &\cmark  & 25.7 & 55.7 & 1.03 & 2.96 & 0.11 & 0.16 \\
FCR + AugMix &\cmark   & 25.3 & 56.2 & 1.02 & 3.00 & 0.11 & 0.16 \\
FCR + TrivAug & \cmark & 25.1 & 54.8 & 1.01 & 2.89 & 0.11 & 0.15 \\
FCR + MixUp &\cmark    & 26.2 & 55.4 & 1.21 & 3.06 & 0.18 & 0.19 \\
FCR + CutMix &\cmark   & 25.4 & 58.0 & 1.19 & 3.26 & 0.19 & 0.19 \\
\bottomrule
\end{tabular}
}
\end{center}
\end{table}

In \cref{tab:imagenet_clean}, we report performance metrics on clean validation images and on the collective of all corrupted validation images from the ImageNet-C collection; the error metrics are averaged across all 15 corruption types and 5 intensity levels. 
We report a breakdown of the error metrics on the ImageNet-C validation images in \cref{tab:imagenetC:breakdown} as differences between runs with and without mollification so as to highlight the performance gains offered by mollification for individual corruption types.

It is interesting to note that even in the ImageNet dataset, mollification offers a wide range of significant improvements on error rate across all corruption types, at the expense of a small degradation in performance on clean validation images. 
\cref{tab:imagenetC:breakdown} shows that these gains are particularly significant for corruptions in the noise and blur categories, as expected

Further investigation could explore hyper-parameters to improve performance overall and to test whether extended training times yield additional performance gains.
Additionally, replicating the experimental setup as for the other datasets, where models are trained from scratch, would provide valuable insights, at the cost of an enormous amount of computing hours.

\section{CONCLUSIONS}
\label{sec:discussion}

We proposed data mollification for supervised image recognition by coupling input mollification and label smoothing to enhance robustness against test-time corruptions. Our work provides a probabilistic interpretation of data mollification and draws connections with augmentations. An interesting future work lies in class-specific noise structures, as well as in the analysis of adversarial robustness. Another future research line is to formally connect mollification with regularization.

There are two additional aspects where we hoped our work could give some breakthrough.
\textbf{(1)} 
We lower bound the augmented likelihood in \cref{eq:auglik} through Jensen's inequality, similarly to \citet{wenzel20posterior}. 
We attempted an unbiased correction of the log-expectation using ideas from \citet{durbin1997}.
\textbf{(2)} 
We employ a likelihood derived from the cross-entropy loss (\cref{eq:crossentropy_likelihood}) and we were able to derive a normalization of this improper likelihood when labels are continuous in $[0,1]$. 
Surprisingly, none of these ways to recover a proper treatment of the likelihood led to improvements, while complicating the implementation and the numerical stability of the training (see derivations in \cref{sec:inference}).  

\subsection*{Acknowledgments}

We acknowledge funding from the Research Council of Finland (grant 334600) and support from the Aalto Science-IT project, the Research Council of Norway (grant 309439 and 315029), CSC, and LUMI.

\newpage

\bibliography{refs}

\begin{thebibliography}{55}
\providecommand{\natexlab}[1]{#1}
\providecommand{\url}[1]{\texttt{#1}}
\expandafter\ifx\csname urlstyle\endcsname\relax
  \providecommand{\doi}[1]{doi: #1}\else
  \providecommand{\doi}{doi: \begingroup \urlstyle{rm}\Url}\fi

\bibitem[Bachmann et~al.(2022)Bachmann, Noci, and Hofmann]{bachmann22tempering}
G.~Bachmann, L.~Noci, and T.~Hofmann.
\newblock How tempering fixes data augmentation in {B}ayesian neural networks.
\newblock In \emph{ICML}, 2022.

\bibitem[Bansal et~al.(2024)Bansal, Borgnia, Chu, Li, Kazemi, Huang, Goldblum, Geiping, and Goldstein]{bansal2024cold}
A.~Bansal, E.~Borgnia, H.-M. Chu, J.~Li, H.~Kazemi, F.~Huang, M.~Goldblum, J.~Geiping, and T.~Goldstein.
\newblock Cold diffusion: Inverting arbitrary image transforms without noise.
\newblock \emph{Advances in Neural Information Processing Systems}, 36, 2024.

\bibitem[Burda et~al.(2016)Burda, Grosse, and Salakhutdinov]{burda2015importance}
Y.~Burda, R.~Grosse, and R.~Salakhutdinov.
\newblock Importance weighted autoencoders.
\newblock In \emph{ICLR}, 2016.

\bibitem[Cifarelli and Regazzini(1996)]{cifarelli1996finetti}
D.~M. Cifarelli and E.~Regazzini.
\newblock De {F}inetti's contribution to probability and statistics.
\newblock \emph{Statistical Science}, 11\penalty0 (4):\penalty0 253--282, 1996.

\bibitem[Cubuk et~al.(2019)Cubuk, Zoph, Man{\'{e}}, Vasudevan, and Le]{cubuk2019autoaugment}
E.~D. Cubuk, B.~Zoph, D.~Man{\'{e}}, V.~Vasudevan, and Q.~V. Le.
\newblock {AutoAugment}: Learning augmentation strategies from data.
\newblock In \emph{CVPR}, 2019.

\bibitem[DeVries and Taylor(2017)]{devries2017cutout}
T.~DeVries and G.~W. Taylor.
\newblock Improved regularization of convolutional neural networks with cutout.
\newblock \emph{arXiv}, 2017.

\bibitem[Durbin and Koopman(1997)]{durbin1997}
J.~Durbin and S.~Koopman.
\newblock Monte {C}arlo maximum likelihood estimation for non-{G}aussian state space models.
\newblock \emph{Biometrika}, 1997.

\bibitem[Fong et~al.(2023)Fong, Holmes, and Walker]{fong2023martingale}
E.~Fong, C.~Holmes, and S.~G. Walker.
\newblock Martingale posterior distributions.
\newblock \emph{Journal of the Royal Statistical Society Series B: Statistical Methodology}, 85\penalty0 (5):\penalty0 1357--1391, 2023.

\bibitem[Guo et~al.(2017)Guo, Pleiss, Sun, and Weinberger]{guo2017calibration}
C.~Guo, G.~Pleiss, Y.~Sun, and K.~Q. Weinberger.
\newblock On calibration of modern neural networks.
\newblock In \emph{ICML}, 2017.

\bibitem[He et~al.(2016)He, Zhang, Ren, and Sun]{he2016identity}
K.~He, X.~Zhang, S.~Ren, and J.~Sun.
\newblock Identity mappings in deep residual networks.
\newblock In \emph{ECCV}, 2016.

\bibitem[Hendrycks and Dietterich(2019)]{hendrycks2019benchmarking}
D.~Hendrycks and T.~Dietterich.
\newblock Benchmarking neural network robustness to common corruptions and perturbations.
\newblock In \emph{ICLR}, 2019.

\bibitem[Hendrycks et~al.(2020)Hendrycks, Mu, Cubuk, Zoph, Gilmer, and Lakshminarayanan]{hendrycks2019augmix}
D.~Hendrycks, N.~Mu, E.~D. Cubuk, B.~Zoph, J.~Gilmer, and B.~Lakshminarayanan.
\newblock Augmix: A simple data processing method to improve robustness and uncertainty.
\newblock In \emph{ICLR}, 2020.

\bibitem[Hendrycks et~al.(2021)Hendrycks, Basart, Mu, Kadavath, Wang, Dorundo, Desai, Zhu, Parajuli, Guo, et~al.]{hendrycks2021many}
D.~Hendrycks, S.~Basart, N.~Mu, S.~Kadavath, F.~Wang, E.~Dorundo, R.~Desai, T.~Zhu, S.~Parajuli, M.~Guo, et~al.
\newblock The many faces of robustness: A critical analysis of out-of-distribution generalization.
\newblock In \emph{ICCV}, 2021.

\bibitem[Hendrycks et~al.(2022)Hendrycks, Zou, Mazeika, Tang, Li, Song, and Steinhardt]{hendrycks22}
D.~Hendrycks, A.~Zou, M.~Mazeika, L.~Tang, B.~Li, D.~Song, and J.~Steinhardt.
\newblock Pix{M}ix: {D}reamlike pictures comprehensively improve safety measures.
\newblock In \emph{CVPR}, 2022.

\bibitem[Ho and Salimans(2021)]{ho2022classifier}
J.~Ho and T.~Salimans.
\newblock Classifier-free diffusion guidance.
\newblock In \emph{NeurIPS workshop on Deep Generative Models}, 2021.

\bibitem[Ho et~al.(2020)Ho, Jain, and Abbeel]{ho2020denoising}
J.~Ho, A.~Jain, and P.~Abbeel.
\newblock Denoising diffusion probabilistic models.
\newblock In \emph{NeurIPS}, 2020.

\bibitem[Hoffer et~al.(2020)Hoffer, Ben-Nun, Hubara, Giladi, Hoefler, and Soudry]{hoffer2020augment}
E.~Hoffer, T.~Ben-Nun, I.~Hubara, N.~Giladi, T.~Hoefler, and D.~Soudry.
\newblock Augment your batch: Improving generalization through instance repetition.
\newblock In \emph{CVPR}, 2020.

\bibitem[Hoogeboom and Salimans(2023)]{hoogeboom2022blurring}
E.~Hoogeboom and T.~Salimans.
\newblock Blurring diffusion models.
\newblock In \emph{ICLR}, 2023.

\bibitem[Izmailov et~al.(2021)Izmailov, Vikram, Hoffman, and Wilson]{izmailov21posterior}
P.~Izmailov, S.~Vikram, M.~D. Hoffman, and A.~G. Wilson.
\newblock What are {B}ayesian neural network posteriors really like?
\newblock In \emph{ICML}, 2021.

\bibitem[Kapoor et~al.(2022)Kapoor, Maddox, Izmailov, and Wilson]{kapoor2022uncertainty}
S.~Kapoor, W.~Maddox, P.~Izmailov, and A.~G. Wilson.
\newblock On uncertainty, tempering, and data augmentation in {B}ayesian classification.
\newblock In \emph{NeurIPS}, 2022.

\bibitem[Kingma et~al.(2021)Kingma, Salimans, Poole, and Ho]{kingma2021variational}
D.~Kingma, T.~Salimans, B.~Poole, and J.~Ho.
\newblock Variational diffusion models.
\newblock In \emph{NeurIPS}, 2021.

\bibitem[Kirichenko et~al.(2023)Kirichenko, Ibrahim, Balestriero, Bouchacourt, Vedantam, Firooz, and Wilson]{Kirichenko2023understanding}
P.~Kirichenko, M.~Ibrahim, R.~Balestriero, D.~Bouchacourt, R.~Vedantam, H.~Firooz, and A.~G. Wilson.
\newblock Understanding the detrimental class-level effects of data augmentation.
\newblock In \emph{NeurIPS}, 2023.

\bibitem[Krizhevsky(2009)]{krizhevsky2009cifar}
A.~Krizhevsky.
\newblock Learning multiple layers of features from tiny images.
\newblock Technical report, 2009.

\bibitem[Le and Yang(2015)]{Le2015TinyIV}
Y.~Le and X.~S. Yang.
\newblock Tiny {ImageNet} visual recognition challenge.
\newblock 2015.

\bibitem[Lee et~al.(2020)Lee, Won, Lee, Lee, Gu, and Hong]{lee2020compounding}
J.~Lee, T.~Won, T.~K. Lee, H.~Lee, G.~Gu, and K.~Hong.
\newblock Compounding the performance improvements of assembled techniques in a convolutional neural network.
\newblock \emph{arXiv}, 2020.

\bibitem[Li et~al.(2023)Li, Prabhudesai, Duggal, Brown, and Pathak]{li2023diffusion}
A.~C. Li, M.~Prabhudesai, S.~Duggal, E.~Brown, and D.~Pathak.
\newblock Your diffusion model is secretly a zero-shot classifier.
\newblock In \emph{ICCV}, 2023.

\bibitem[Li et~al.(2020)Li, Dasarathy, and Berisha]{li20structured}
W.~Li, G.~Dasarathy, and V.~Berisha.
\newblock Regularization via structural label smoothing.
\newblock In \emph{AISTATS}, 2020.

\bibitem[Lienen and Hüllermeier(2021)]{lienen21labelrelax}
J.~Lienen and E.~Hüllermeier.
\newblock From label smoothing to label relaxation.
\newblock In \emph{AAAI}, 2021.

\bibitem[Loshchilov and Hutter(2016)]{loshchilov2016sgdr}
I.~Loshchilov and F.~Hutter.
\newblock Sgdr: Stochastic gradient descent with warm restarts.
\newblock In \emph{ICLR}, 2016.

\bibitem[Luo et~al.(2020)Luo, Beatson, Norouzi, Zhu, Duvenaud, Adams, and Chen]{luo2020sumo}
Y.~Luo, A.~Beatson, M.~Norouzi, J.~Zhu, D.~Duvenaud, R.~P. Adams, and R.~T. Chen.
\newblock Sumo: Unbiased estimation of log marginal probability for latent variable models.
\newblock In \emph{ICLR}, 2020.

\bibitem[Maher and Kull(2021)]{maher21instance}
M.~Maher and M.~Kull.
\newblock Instance-based label smoothing for better calibrated classification networks.
\newblock In \emph{ICMLA}, 2021.

\bibitem[M{\"u}ller et~al.(2019)M{\"u}ller, Kornblith, and Hinton]{muller2019does}
R.~M{\"u}ller, S.~Kornblith, and G.~E. Hinton.
\newblock When does label smoothing help?
\newblock In \emph{NeurIPS}, 2019.

\bibitem[M{\"u}ller and Hutter(2021)]{muller2021trivialaugment}
S.~G. M{\"u}ller and F.~Hutter.
\newblock Trivialaugment: Tuning-free yet state-of-the-art data augmentation.
\newblock In \emph{ICCV}, 2021.

\bibitem[Nabarro et~al.(2022)Nabarro, Ganev, Garriga-Alonso, Fortuin, van~der Wilk, and Aitchison]{nabarro22augmentation}
S.~Nabarro, S.~Ganev, A.~Garriga-Alonso, V.~Fortuin, M.~van~der Wilk, and L.~Aitchison.
\newblock Data augmentation in {B}ayesian neural networks and the cold posterior effect.
\newblock In \emph{UAI}, 2022.

\bibitem[Nichol and Dhariwal(2021)]{pmlr-v139-nichol21a}
A.~Nichol and P.~Dhariwal.
\newblock Improved denoising diffusion probabilistic models.
\newblock In \emph{ICML}, 2021.

\bibitem[Qin et~al.(2023)Qin, Wang, Lakshminarayanan, Chi, and Beutel]{quin2023autolabel}
Y.~Qin, X.~Wang, B.~Lakshminarayanan, E.~H. Chi, and A.~Beutel.
\newblock What are effective labels for augmented data? {I}mproving calibration and robustness with {AutoLabel}.
\newblock In \emph{IEEE Conference on Secure and Trustworthy Machine Learning}, 2023.

\bibitem[Radford et~al.(2021)Radford, Kim, Hallacy, Ramesh, Goh, Agarwal, Sastry, Askell, Mishkin, Clark, et~al.]{radford2021learning}
A.~Radford, J.~W. Kim, C.~Hallacy, A.~Ramesh, G.~Goh, S.~Agarwal, G.~Sastry, A.~Askell, P.~Mishkin, J.~Clark, et~al.
\newblock Learning transferable visual models from natural language supervision.
\newblock In \emph{ICML}, 2021.

\bibitem[Rissanen et~al.(2023)Rissanen, Heinonen, and Solin]{rissanen2022generative}
S.~Rissanen, M.~Heinonen, and A.~Solin.
\newblock Generative modelling with inverse heat dissipation.
\newblock In \emph{ICLR}, 2023.

\bibitem[Schneider et~al.(2020)Schneider, Rusak, Eck, Bringmann, Brendel, and Bethge]{schneider2020improving}
S.~Schneider, E.~Rusak, L.~Eck, O.~Bringmann, W.~Brendel, and M.~Bethge.
\newblock Improving robustness against common corruptions by covariate shift adaptation.
\newblock In \emph{NeurIPS}, 2020.

\bibitem[Sensoy et~al.(2018)Sensoy, Kaplan, and Kandemir]{sensoy2018evidential}
M.~Sensoy, L.~Kaplan, and M.~Kandemir.
\newblock Evidential deep learning to quantify classification uncertainty.
\newblock In \emph{NeurIPS}, 2018.

\bibitem[Shephard and Pitt(1997)]{shephard1997}
N.~Shephard and M.~Pitt.
\newblock Likelihood analysis of non-{G}aussian measurement time series.
\newblock \emph{Biometrika}, 1997.

\bibitem[Song et~al.(2021)Song, Sohl-Dickstein, Kingma, Kumar, Ermon, and Poole]{song2020score}
Y.~Song, J.~Sohl-Dickstein, D.~P. Kingma, A.~Kumar, S.~Ermon, and B.~Poole.
\newblock Score-based generative modeling through stochastic differential equations.
\newblock In \emph{ICLR}, 2021.

\bibitem[Szegedy et~al.(2016)Szegedy, Vanhoucke, Ioffe, Shlens, and Wojna]{szegedy2016rethinking}
C.~Szegedy, V.~Vanhoucke, S.~Ioffe, J.~Shlens, and Z.~Wojna.
\newblock Rethinking the inception architecture for computer vision.
\newblock In \emph{CVPR}, 2016.

\bibitem[Teney et~al.(2024)Teney, Wang, and Abbasnejad]{teney2024selective}
D.~Teney, J.~Wang, and E.~Abbasnejad.
\newblock Selective mixup helps with distribution shifts, but not (only) because of mixup.
\newblock In \emph{ICML}, 2024.

\bibitem[Thulasidasan et~al.(2019)Thulasidasan, Chennupati, Bilmes, Bhattacharya, and Michalak]{thulasidasan2019calibration}
S.~Thulasidasan, G.~Chennupati, J.~A. Bilmes, T.~Bhattacharya, and S.~Michalak.
\newblock On mixup training: Improved calibration and predictive uncertainty for deep neural networks.
\newblock In \emph{NeurIPS}, 2019.

\bibitem[Tran et~al.(2023)Tran, Franzese, Michiardi, and Filippone]{tran2023}
B.-H. Tran, G.~Franzese, P.~Michiardi, and M.~Filippone.
\newblock One-line-of-code data mollification improves optimization of likelihood-based generative models.
\newblock In \emph{NeurIPS}, 2023.

\bibitem[Vryniotis(2021)]{torchvision}
V.~Vryniotis.
\newblock {How to Train State-Of-The-Art Models Using TorchVision’s Latest Primitives}.
\newblock \url{pytorch.org}, 2021.

\bibitem[Wang et~al.(2023)Wang, Polson, and Sokolov]{wang23BA}
Y.~Wang, N.~Polson, and V.~O. Sokolov.
\newblock {Data Augmentation for Bayesian Deep Learning}.
\newblock \emph{Bayesian Analysis}, 2023.

\bibitem[Wenzel et~al.(2020)Wenzel, Roth, Veeling, Swiatkowski, Tran, Mandt, Snoek, Salimans, Jenatton, and Nowozin]{wenzel20posterior}
F.~Wenzel, K.~Roth, B.~Veeling, J.~Swiatkowski, L.~Tran, S.~Mandt, J.~Snoek, T.~Salimans, R.~Jenatton, and S.~Nowozin.
\newblock How good is the {B}ayes posterior in deep neural networks really?
\newblock In \emph{ICML}, 2020.

\bibitem[Wightman et~al.(2021)Wightman, Touvron, and J{\'e}gou]{wightman2021resnet}
R.~Wightman, H.~Touvron, and H.~J{\'e}gou.
\newblock Resnet strikes back: An improved training procedure in timm.
\newblock In \emph{NeurIPS workshop on ImageNet PPF}, 2021.

\bibitem[Wu and Williamson(2024)]{wu24}
L.~Wu and S.~A. Williamson.
\newblock Posterior uncertainty quantification in neural networks using data augmentation.
\newblock In \emph{AISTATS}, 2024.

\bibitem[Yao et~al.(2022)Yao, Wang, Li, Zhang, Liang, Zou, and Finn]{yao2022improving}
H.~Yao, Y.~Wang, S.~Li, L.~Zhang, W.~Liang, J.~Zou, and C.~Finn.
\newblock Improving out-of-distribution robustness via selective augmentation.
\newblock In \emph{ICML}, 2022.

\bibitem[Yun et~al.(2019)Yun, Han, Oh, Chun, Choe, and Yoo]{yun2019cutmix}
S.~Yun, D.~Han, S.~J. Oh, S.~Chun, J.~Choe, and Y.~Yoo.
\newblock Cutmix: Regularization strategy to train strong classifiers with localizable features.
\newblock In \emph{ICCV}, 2019.

\bibitem[Zhang et~al.(2018)Zhang, Cisse, Dauphin, and Lopez-Paz]{zhang2017mixup}
H.~Zhang, M.~Cisse, Y.~N. Dauphin, and D.~Lopez-Paz.
\newblock mixup: Beyond empirical risk minimization.
\newblock In \emph{ICLR}, 2018.

\bibitem[Zhang et~al.(2022)Zhang, Levine, and Finn]{zhang2022memo}
M.~Zhang, S.~Levine, and C.~Finn.
\newblock Memo: Test time robustness via adaptation and augmentation.
\newblock In \emph{NeurIPS}, 2022.

\end{thebibliography}

\section*{Checklist}

 \begin{enumerate}

 \item For all models and algorithms presented, check if you include:
 \begin{enumerate}
   \item A clear description of the mathematical setting, assumptions, algorithm, and/or model. [Yes]
   \item An analysis of the properties and complexity (time, space, sample size) of any algorithm. [Not Applicable]
   \item (Optional) Anonymized source code, with specification of all dependencies, including external libraries. [Yes]
 \end{enumerate}

 \item For any theoretical claim, check if you include:
 \begin{enumerate}
   \item Statements of the full set of assumptions of all theoretical results. [Yes]
   \item Complete proofs of all theoretical results. [Yes]
   \item Clear explanations of any assumptions. [Yes]     
 \end{enumerate}

 \item For all figures and tables that present empirical results, check if you include:
 \begin{enumerate}
   \item The code, data, and instructions needed to reproduce the main experimental results (either in the supplemental material or as a URL). [Yes]
   \item All the training details (e.g., data splits, hyperparameters, how they were chosen). [Yes]
         \item A clear definition of the specific measure or statistics and error bars (e.g., with respect to the random seed after running experiments multiple times). [Yes]
         \item A description of the computing infrastructure used. (e.g., type of GPUs, internal cluster, or cloud provider). [Yes]
 \end{enumerate}

 \item If you are using existing assets (e.g., code, data, models) or curating/releasing new assets, check if you include:
 \begin{enumerate}
   \item Citations of the creator If your work uses existing assets. [Yes]
   \item The license information of the assets, if applicable. [Not Applicable]
   \item New assets either in the supplemental material or as a URL, if applicable. [Not Applicable]
   \item Information about consent from data providers/curators. [Not Applicable]
   \item Discussion of sensible content if applicable, e.g., personally identifiable information or offensive content. [Not Applicable]
 \end{enumerate}

 \item If you used crowdsourcing or conducted research with human subjects, check if you include:
 \begin{enumerate}
   \item The full text of instructions given to participants and screenshots. [Not Applicable]
   \item Descriptions of potential participant risks, with links to Institutional Review Board (IRB) approvals if applicable. [Not Applicable]
   \item The estimated hourly wage paid to participants and the total amount spent on participant compensation. [Not Applicable]
 \end{enumerate}

 \end{enumerate}

\clearpage
\newpage

\onecolumn

\appendix

\section{ADDITIONAL RESULTS}

\subsection{Result table from the main paper with standard deviations}

For completeness, we report Table 1 from the main paper where we add standard deviations for all metrics. 

\begin{table}[h]
\begin{center}
\caption{Results on CIFAR-10, CIFAR-100, and TinyImageNet with standard deviations in parentheses.}
\label{tab:benchmark3}
\resizebox{0.99\linewidth}{!}{
\setlength{\tabcolsep}{4pt}
\begin{tabular}{lc cccccc cccccc cccccc}
& & \multicolumn{6}{c}{CIFAR-10} & \multicolumn{6}{c}{CIFAR-100} & \multicolumn{6}{c}{TinyImageNet} \\
\cmidrule(lr){3-8} \cmidrule(lr){9-14} \cmidrule(lr){15-20}
& & \multicolumn{2}{c}{Error $(\downarrow)$} & \multicolumn{2}{c}{NLL $(\downarrow)$} & \multicolumn{2}{c}{ECE $(\downarrow)$} & \multicolumn{2}{c}{Error} & \multicolumn{2}{c}{NLL} & \multicolumn{2}{c}{ECE} & \multicolumn{2}{c}{Error} & \multicolumn{2}{c}{NLL} & \multicolumn{2}{c}{ECE} \\
\cmidrule(lr){3-4} \cmidrule(lr){5-6} \cmidrule(lr){7-8} \cmidrule(lr){9-10} \cmidrule(lr){11-12} \cmidrule(lr){13-14} \cmidrule(lr){15-16} \cmidrule(lr){17-18} \cmidrule(lr){19-20}
\rowcolor{white} Augmentation & Moll. & clean & corr & clean & corr & clean & corr & clean & corr & clean & corr & clean & corr & clean & corr & clean & corr & clean & corr \\
\toprule
presnet50 & \xmark & 11.7 & 29.4 & 0.49 & 1.34 & 0.09 & 0.22 & 39.2 & 60.3 & 1.64 & 2.89 & 0.16 & 0.24 & 57.3 & 84.6 & 2.73 & 4.30 & 0.16 & 0.11 \\ 
 & & \textcolor{gray}{(0.10)} & \textcolor{gray}{(0.63)} & \textcolor{gray}{(0.01)} & \textcolor{gray}{(0.04)} & \textcolor{gray}{(0.00)} & \textcolor{gray}{(0.01)} & \textcolor{gray}{(0.40)} & \textcolor{gray}{(0.39)} & \textcolor{gray}{(0.02)} & \textcolor{gray}{(0.06)} & \textcolor{gray}{(0.01)} & \textcolor{gray}{(0.01)} & \textcolor{gray}{(2.12)} & \textcolor{gray}{(0.33)} & \textcolor{gray}{(0.13)} & \textcolor{gray}{(0.02)} & \textcolor{gray}{(0.01)} & \textcolor{gray}{(0.00)} \\ 
FCR & \xmark & 4.9 & 21.2 & 0.22 & 1.14 & 0.04 & 0.17 & 23.2 & 47.1 & 0.99 & 2.32 & 0.12 & 0.21 & 33.9 & 76.0 & 1.53 & 4.11 & 0.12 & 0.21 \\ 
 & & \textcolor{gray}{(0.05)} & \textcolor{gray}{(1.01)} & \textcolor{gray}{(0.00)} & \textcolor{gray}{(0.15)} & \textcolor{gray}{(0.00)} & \textcolor{gray}{(0.01)} & \textcolor{gray}{(0.19)} & \textcolor{gray}{(0.33)} & \textcolor{gray}{(0.01)} & \textcolor{gray}{(0.06)} & \textcolor{gray}{(0.00)} & \textcolor{gray}{(0.01)} & \textcolor{gray}{(0.31)} & \textcolor{gray}{(0.16)} & \textcolor{gray}{(0.01)} & \textcolor{gray}{(0.02)} & \textcolor{gray}{(0.00)} & \textcolor{gray}{(0.00)} \\ 
FCR+RandAug & \xmark & 4.2 & 15.2 & 0.18 & 0.73 & 0.04 & 0.12 & 21.4 & 41.4 & 0.87 & 2.09 & 0.11 & 0.19 & 32.4 & 70.4 & 1.45 & 3.88 & 0.12 & 0.22 \\ 
 & & \textcolor{gray}{(0.15)} & \textcolor{gray}{(0.41)} & \textcolor{gray}{(0.00)} & \textcolor{gray}{(0.03)} & \textcolor{gray}{(0.00)} & \textcolor{gray}{(0.00)} & \textcolor{gray}{(0.15)} & \textcolor{gray}{(0.59)} & \textcolor{gray}{(0.00)} & \textcolor{gray}{(0.07)} & \textcolor{gray}{(0.00)} & \textcolor{gray}{(0.00)} & \textcolor{gray}{(0.15)} & \textcolor{gray}{(0.20)} & \textcolor{gray}{(0.01)} & \textcolor{gray}{(0.04)} & \textcolor{gray}{(0.00)} & \textcolor{gray}{(0.00)} \\ 
FCR+AutoAug & \xmark & 4.6 & 13.3 & 0.19 & 0.60 & 0.04 & 0.10 & 22.7 & 38.8 & 0.91 & 1.86 & 0.11 & 0.17 & 33.8 & 69.5 & 1.48 & 3.77 & 0.11 & 0.21 \\ 
 & & \textcolor{gray}{(0.07)} & \textcolor{gray}{(0.47)} & \textcolor{gray}{(0.00)} & \textcolor{gray}{(0.04)} & \textcolor{gray}{(0.00)} & \textcolor{gray}{(0.00)} & \textcolor{gray}{(0.33)} & \textcolor{gray}{(0.36)} & \textcolor{gray}{(0.01)} & \textcolor{gray}{(0.05)} & \textcolor{gray}{(0.00)} & \textcolor{gray}{(0.00)} & \textcolor{gray}{(0.29)} & \textcolor{gray}{(0.24)} & \textcolor{gray}{(0.01)} & \textcolor{gray}{(0.05)} & \textcolor{gray}{(0.00)} & \textcolor{gray}{(0.00)} \\ 
FCR+AugMix & \xmark & 4.4 & 10.7 & 0.18 & 0.43 & 0.04 & 0.08 & 23.0 & 35.5 & 0.94 & 1.60 & 0.12 & 0.16 & 34.0 & 63.5 & 1.52 & 3.37 & 0.11 & 0.20 \\ 
 & & \textcolor{gray}{(0.06)} & \textcolor{gray}{(0.26)} & \textcolor{gray}{(0.00)} & \textcolor{gray}{(0.01)} & \textcolor{gray}{(0.00)} & \textcolor{gray}{(0.00)} & \textcolor{gray}{(0.31)} & \textcolor{gray}{(0.34)} & \textcolor{gray}{(0.02)} & \textcolor{gray}{(0.02)} & \textcolor{gray}{(0.00)} & \textcolor{gray}{(0.00)} & \textcolor{gray}{(0.28)} & \textcolor{gray}{(0.18)} & \textcolor{gray}{(0.01)} & \textcolor{gray}{(0.03)} & \textcolor{gray}{(0.00)} & \textcolor{gray}{(0.01)} \\ 
FCR+TrivAug & \xmark & 3.4 & 12.2 & 0.12 & 0.50 & 0.03 & 0.09 & 20.2 & 36.8 & 0.78 & 1.69 & 0.10 & 0.16 & 33.8 & 69.9 & 1.57 & 4.64 & 0.13 & 0.30 \\ 
 & & \textcolor{gray}{(0.02)} & \textcolor{gray}{(0.37)} & \textcolor{gray}{(0.00)} & \textcolor{gray}{(0.02)} & \textcolor{gray}{(0.00)} & \textcolor{gray}{(0.00)} & \textcolor{gray}{(0.17)} & \textcolor{gray}{(0.50)} & \textcolor{gray}{(0.01)} & \textcolor{gray}{(0.05)} & \textcolor{gray}{(0.00)} & \textcolor{gray}{(0.00)} & \textcolor{gray}{(0.13)} & \textcolor{gray}{(0.35)} & \textcolor{gray}{(0.01)} & \textcolor{gray}{(0.12)} & \textcolor{gray}{(0.00)} & \textcolor{gray}{(0.00)} \\ 
FCR+MixUp & \xmark & 3.8 & 21.9 & 0.29 & 0.83 & 0.17 & 0.20 & 21.3 & 45.1 & 0.96 & 2.06 & 0.15 & 0.18 & 33.3 & 72.5 & 1.75 & 3.67 & 0.22 & 0.15 \\ 
 & & \textcolor{gray}{(0.17)} & \textcolor{gray}{(1.47)} & \textcolor{gray}{(0.02)} & \textcolor{gray}{(0.04)} & \textcolor{gray}{(0.01)} & \textcolor{gray}{(0.02)} & \textcolor{gray}{(0.49)} & \textcolor{gray}{(0.72)} & \textcolor{gray}{(0.04)} & \textcolor{gray}{(0.06)} & \textcolor{gray}{(0.02)} & \textcolor{gray}{(0.03)} & \textcolor{gray}{(0.36)} & \textcolor{gray}{(0.35)} & \textcolor{gray}{(0.04)} & \textcolor{gray}{(0.03)} & \textcolor{gray}{(0.01)} & \textcolor{gray}{(0.01)} \\ 
 FCR+CutMix & \xmark & 3.9 & 28.2 & 0.22 & 1.19 & 0.11 & 0.19 & 22.6 & 54.3 & 1.12 & 2.81 & 0.18 & 0.22 & 31.7 & 76.4 & 1.52 & 4.05 & 0.13 & 0.18 \\ 
 & & \textcolor{gray}{(0.17)} & \textcolor{gray}{(0.73)} & \textcolor{gray}{(0.01)} & \textcolor{gray}{(0.11)} & \textcolor{gray}{(0.01)} & \textcolor{gray}{(0.02)} & \textcolor{gray}{(0.05)} & \textcolor{gray}{(0.36)} & \textcolor{gray}{(0.00)} & \textcolor{gray}{(0.11)} & \textcolor{gray}{(0.01)} & \textcolor{gray}{(0.02)} & \textcolor{gray}{(0.27)} & \textcolor{gray}{(0.21)} & \textcolor{gray}{(0.02)} & \textcolor{gray}{(0.04)} & \textcolor{gray}{(0.00)} & \textcolor{gray}{(0.01)} \\ 
\midrule 
presnet50 & \cmark & 15.3 & 20.6 & 0.65 & 0.83 & 0.11 & 0.13 & 49.5 & 54.7 & 2.29 & 2.54 & 0.15 & 0.14 & 50.1 & 67.3 & 2.42 & 3.39 & 0.15 & 0.14 \\ 
 & & \textcolor{gray}{(0.07)} & \textcolor{gray}{(0.21)} & \textcolor{gray}{(0.01)} & \textcolor{gray}{(0.01)} & \textcolor{gray}{(0.00)} & \textcolor{gray}{(0.00)} & \textcolor{gray}{(0.20)} & \textcolor{gray}{(0.17)} & \textcolor{gray}{(0.01)} & \textcolor{gray}{(0.01)} & \textcolor{gray}{(0.00)} & \textcolor{gray}{(0.00)} & \textcolor{gray}{(0.18)} & \textcolor{gray}{(0.19)} & \textcolor{gray}{(0.01)} & \textcolor{gray}{(0.01)} & \textcolor{gray}{(0.00)} & \textcolor{gray}{(0.00)} \\ 
FCR & \cmark & 5.8 & 10.2 & 0.24 & 0.43 & 0.04 & 0.08 & 27.0 & 34.7 & 1.16 & 1.56 & 0.11 & 0.13 & 35.3 & 60.1 & 1.63 & 3.04 & 0.12 & 0.15 \\ 
 & & \textcolor{gray}{(0.17)} & \textcolor{gray}{(0.15)} & \textcolor{gray}{(0.01)} & \textcolor{gray}{(0.01)} & \textcolor{gray}{(0.00)} & \textcolor{gray}{(0.00)} & \textcolor{gray}{(0.07)} & \textcolor{gray}{(0.19)} & \textcolor{gray}{(0.01)} & \textcolor{gray}{(0.01)} & \textcolor{gray}{(0.00)} & \textcolor{gray}{(0.00)} & \textcolor{gray}{(0.07)} & \textcolor{gray}{(0.20)} & \textcolor{gray}{(0.01)} & \textcolor{gray}{(0.02)} & \textcolor{gray}{(0.00)} & \textcolor{gray}{(0.01)} \\ 
FCR+RandAug & \cmark & 4.4 & 8.3 & 0.18 & 0.34 & 0.04 & 0.07 & 23.2 & 30.6 & 0.96 & 1.33 & 0.10 & 0.13 & 33.5 & 57.2 & 1.51 & 2.85 & 0.11 & 0.15 \\ 
 & & \textcolor{gray}{(0.06)} & \textcolor{gray}{(0.10)} & \textcolor{gray}{(0.00)} & \textcolor{gray}{(0.01)} & \textcolor{gray}{(0.00)} & \textcolor{gray}{(0.00)} & \textcolor{gray}{(0.07)} & \textcolor{gray}{(0.18)} & \textcolor{gray}{(0.00)} & \textcolor{gray}{(0.01)} & \textcolor{gray}{(0.00)} & \textcolor{gray}{(0.00)} & \textcolor{gray}{(0.27)} & \textcolor{gray}{(0.16)} & \textcolor{gray}{(0.01)} & \textcolor{gray}{(0.01)} & \textcolor{gray}{(0.00)} & \textcolor{gray}{(0.00)} \\ 
FCR+AutoAug & \cmark & 4.2 & 8.2 & 0.17 & 0.33 & 0.04 & 0.07 & 23.7 & 30.8 & 0.96 & 1.32 & 0.10 & 0.13 & 34.1 & 58.0 & 1.52 & 2.90 & 0.11 & 0.15 \\ 
 & & \textcolor{gray}{(0.05)} & \textcolor{gray}{(0.15)} & \textcolor{gray}{(0.00)} & \textcolor{gray}{(0.01)} & \textcolor{gray}{(0.00)} & \textcolor{gray}{(0.00)} & \textcolor{gray}{(0.12)} & \textcolor{gray}{(0.24)} & \textcolor{gray}{(0.00)} & \textcolor{gray}{(0.01)} & \textcolor{gray}{(0.00)} & \textcolor{gray}{(0.00)} & \textcolor{gray}{(0.06)} & \textcolor{gray}{(0.27)} & \textcolor{gray}{(0.01)} & \textcolor{gray}{(0.03)} & \textcolor{gray}{(0.00)} & \textcolor{gray}{(0.00)} \\ 
FCR+AugMix & \cmark & 5.4 & 8.7 & 0.22 & 0.36 & 0.04 & 0.08 & 25.7 & 32.3 & 1.09 & 1.42 & 0.10 & 0.13 & 36.0 & 57.1 & 1.66 & 2.84 & 0.12 & 0.15 \\ 
 & & \textcolor{gray}{(0.11)} & \textcolor{gray}{(0.11)} & \textcolor{gray}{(0.00)} & \textcolor{gray}{(0.01)} & \textcolor{gray}{(0.00)} & \textcolor{gray}{(0.00)} & \textcolor{gray}{(0.30)} & \textcolor{gray}{(0.11)} & \textcolor{gray}{(0.01)} & \textcolor{gray}{(0.00)} & \textcolor{gray}{(0.00)} & \textcolor{gray}{(0.00)} & \textcolor{gray}{(0.27)} & \textcolor{gray}{(0.19)} & \textcolor{gray}{(0.01)} & \textcolor{gray}{(0.01)} & \textcolor{gray}{(0.00)} & \textcolor{gray}{(0.00)} \\ 
FCR+TrivAug & \cmark & 3.7 & 7.7 & 0.14 & 0.30 & 0.03 & 0.07 & 20.5 & 28.4 & 0.81 & 1.19 & 0.09 & 0.12 & 32.4 & 59.2 & 1.43 & 3.04 & 0.11 & 0.17 \\ 
 & & \textcolor{gray}{(0.07)} & \textcolor{gray}{(0.33)} & \textcolor{gray}{(0.00)} & \textcolor{gray}{(0.02)} & \textcolor{gray}{(0.00)} & \textcolor{gray}{(0.00)} & \textcolor{gray}{(0.20)} & \textcolor{gray}{(0.35)} & \textcolor{gray}{(0.01)} & \textcolor{gray}{(0.02)} & \textcolor{gray}{(0.00)} & \textcolor{gray}{(0.00)} & \textcolor{gray}{(0.21)} & \textcolor{gray}{(0.62)} & \textcolor{gray}{(0.01)} & \textcolor{gray}{(0.05)} & \textcolor{gray}{(0.00)} & \textcolor{gray}{(0.00)} \\ 
FCR+MixUp & \cmark & 4.5 & 9.4 & 0.34 & 0.50 & 0.20 & 0.21 & 23.3 & 32.3 & 1.12 & 1.51 & 0.20 & 0.20 & 34.3 & 61.9 & 1.81 & 3.15 & 0.23 & 0.18 \\ 
 & & \textcolor{gray}{(0.10)} & \textcolor{gray}{(0.22)} & \textcolor{gray}{(0.01)} & \textcolor{gray}{(0.01)} & \textcolor{gray}{(0.01)} & \textcolor{gray}{(0.00)} & \textcolor{gray}{(0.38)} & \textcolor{gray}{(0.30)} & \textcolor{gray}{(0.03)} & \textcolor{gray}{(0.01)} & \textcolor{gray}{(0.01)} & \textcolor{gray}{(0.00)} & \textcolor{gray}{(0.29)} & \textcolor{gray}{(0.25)} & \textcolor{gray}{(0.02)} & \textcolor{gray}{(0.03)} & \textcolor{gray}{(0.00)} & \textcolor{gray}{(0.01)} \\ 
 FCR+CutMix & \cmark & 4.0 & 10.7 & 0.26 & 0.48 & 0.15 & 0.17 & 22.5 & 34.7 & 1.13 & 1.66 & 0.21 & 0.19 & 31.2 & 65.0 & 1.65 & 3.39 & 0.20 & 0.17 \\ 
 & & \textcolor{gray}{(0.06)} & \textcolor{gray}{(0.24)} & \textcolor{gray}{(0.01)} & \textcolor{gray}{(0.01)} & \textcolor{gray}{(0.00)} & \textcolor{gray}{(0.00)} & \textcolor{gray}{(0.06)} & \textcolor{gray}{(0.20)} & \textcolor{gray}{(0.03)} & \textcolor{gray}{(0.02)} & \textcolor{gray}{(0.02)} & \textcolor{gray}{(0.01)} & \textcolor{gray}{(0.34)} & \textcolor{gray}{(0.20)} & \textcolor{gray}{(0.03)} & \textcolor{gray}{(0.01)} & \textcolor{gray}{(0.01)} & \textcolor{gray}{(0.00)} \\ 
\bottomrule 
\end{tabular} 
}
\end{center}
\end{table}

\subsection{Neural network architecture comparison}
\label{sec:arch}

\begin{table}[h!]
\begin{center}
\caption{Results on CIFAR-10, CIFAR-100, and TinyImageNet over multiple network architectures. All results using mollification.}
\label{tab:benchmark-arch}
\resizebox{0.99\linewidth}{!}{
\setlength{\tabcolsep}{4pt}
\begin{tabular}{lc cccccc cccccc cccccc}
& & \multicolumn{6}{c}{CIFAR-10} & \multicolumn{6}{c}{CIFAR-100} & \multicolumn{6}{c}{TinyImageNet} \\
\cmidrule(lr){3-8} \cmidrule(lr){9-14} \cmidrule(lr){15-20}
& & \multicolumn{2}{c}{Error $(\downarrow)$} & \multicolumn{2}{c}{NLL $(\downarrow)$} & \multicolumn{2}{c}{ECE $(\downarrow)$} & \multicolumn{2}{c}{Error} & \multicolumn{2}{c}{NLL} & \multicolumn{2}{c}{ECE} & \multicolumn{2}{c}{Error} & \multicolumn{2}{c}{NLL} & \multicolumn{2}{c}{ECE} \\
\cmidrule(lr){3-4} \cmidrule(lr){5-6} \cmidrule(lr){7-8} \cmidrule(lr){9-10} \cmidrule(lr){11-12} \cmidrule(lr){13-14} \cmidrule(lr){15-16} \cmidrule(lr){17-18} \cmidrule(lr){19-20}
\rowcolor{white} Network & Aug. & clean & corr & clean & corr & clean & corr & clean & corr & clean & corr & clean & corr & clean & corr & clean & corr & clean & corr \\
\toprule
AllConvNet-10 &  & 8.8 & 15.3 & 0.28 & 0.52 & 0.06 & 0.10 & 31.1 & 41.1 & 1.25 & 1.74 & 0.14 & 0.16 & 45.6 & 69.3 & 2.00 & 3.47 & 0.15 & 0.18 \\ 
ResNeXt-29 &  & 14.2 & 19.7 & 0.44 & 0.66 & 0.08 & 0.12 & 39.4 & 46.7 & 1.88 & 2.29 & 0.17 & 0.20 & 49.0 & 69.9 & 2.32 & 3.49 & 0.15 & 0.15 \\ 
DenseNet-40 &  & 13.2 & 18.9 & 0.54 & 0.78 & 0.10 & 0.13 & 39.8 & 48.9 & 1.67 & 2.13 & 0.14 & 0.15 & 50.7 & 73.2 & 2.23 & 3.62 & 0.13 & 0.16 \\ 
WRN-40 &  & 15.7 & 20.4 & 0.71 & 0.87 & 0.12 & 0.14 & 44.9 & 51.0 & 2.15 & 2.44 & 0.22 & 0.21 & 51.6 & 69.5 & 2.33 & 3.41 & 0.15 & 0.16 \\ 
GoogleNet &  & 11.0 & 16.0 & 0.44 & 0.63 & 0.08 & 0.11 & 39.9 & 46.3 & 1.84 & 2.18 & 0.14 & 0.14 & 39.9 & 60.1 & 1.93 & 3.10 & 0.14 & 0.16 \\ 
\midrule
AllConvNet-10 & FCR & 7.0 & 13.5 & 0.23 & 0.46 & 0.05 & 0.09 & 29.4 & 39.7 & 1.13 & 1.66 & 0.13 & 0.16 & 42.1 & 68.2 & 1.77 & 3.45 & 0.13 & 0.19 \\ 
ResNeXt-29 & FCR & 5.6 & 11.4 & 0.20 & 0.43 & 0.05 & 0.09 & 29.7 & 38.9 & 1.27 & 1.76 & 0.12 & 0.15 & 39.3 & 68.4 & 1.75 & 3.48 & 0.12 & 0.17 \\ 
DenseNet-40 & FCR & 7.4 & 13.2 & 0.27 & 0.50 & 0.05 & 0.09 & 30.8 & 41.8 & 1.18 & 1.75 & 0.12 & 0.15 & 43.1 & 69.7 & 1.77 & 3.42 & 0.12 & 0.16 \\ 
WRN-40 & FCR & 6.3 & 11.2 & 0.24 & 0.45 & 0.05 & 0.09 & 29.3 & 38.1 & 1.25 & 1.72 & 0.14 & 0.17 & 40.3 & 69.8 & 1.81 & 3.79 & 0.15 & 0.25 \\ 
GoogleNet & FCR & 6.4 & 10.6 & 0.25 & 0.43 & 0.05 & 0.09 & 28.2 & 35.8 & 1.22 & 1.62 & 0.11 & 0.14 & 37.2 & 60.9 & 1.72 & 3.06 & 0.12 & 0.15 \\ 
\midrule
AllConvNet-10 & FCR+AugMix & 7.0 & 12.5 & 0.23 & 0.41 & 0.05 & 0.08 & 29.2 & 38.2 & 1.11 & 1.53 & 0.13 & 0.15 & 41.5 & 64.4 & 1.74 & 3.12 & 0.13 & 0.17 \\ 
ResNeXt-29 & FCR+AugMix & 5.0 & 9.6 & 0.18 & 0.35 & 0.04 & 0.08 & 27.9 & 35.9 & 1.15 & 1.55 & 0.11 & 0.13 & 39.0 & 63.8 & 1.71 & 3.19 & 0.12 & 0.17 \\ 
DenseNet-40 & FCR+AugMix & 7.0 & 12.2 & 0.25 & 0.45 & 0.05 & 0.09 & 29.5 & 39.6 & 1.11 & 1.59 & 0.12 & 0.14 & 42.4 & 66.3 & 1.74 & 3.14 & 0.12 & 0.16 \\ 
WRN-40 & FCR+AugMix & 6.4 & 10.3 & 0.23 & 0.40 & 0.04 & 0.08 & 29.1 & 36.8 & 1.22 & 1.60 & 0.13 & 0.16 & 39.5 & 63.4 & 1.71 & 3.21 & 0.14 & 0.21 \\ 
GoogleNet & FCR+AugMix & 5.5 & 9.2 & 0.22 & 0.37 & 0.04 & 0.08 & 27.7 & 34.2 & 1.18 & 1.50 & 0.11 & 0.13 & 38.2 & 58.5 & 1.74 & 2.89 & 0.13 & 0.16 \\ 
\bottomrule 
\end{tabular} 
}
\end{center}
\end{table}

\subsection{Mollification vs severity}

\begin{figure*}[h!]
    \centering
    \includegraphics[width=0.7\textwidth]{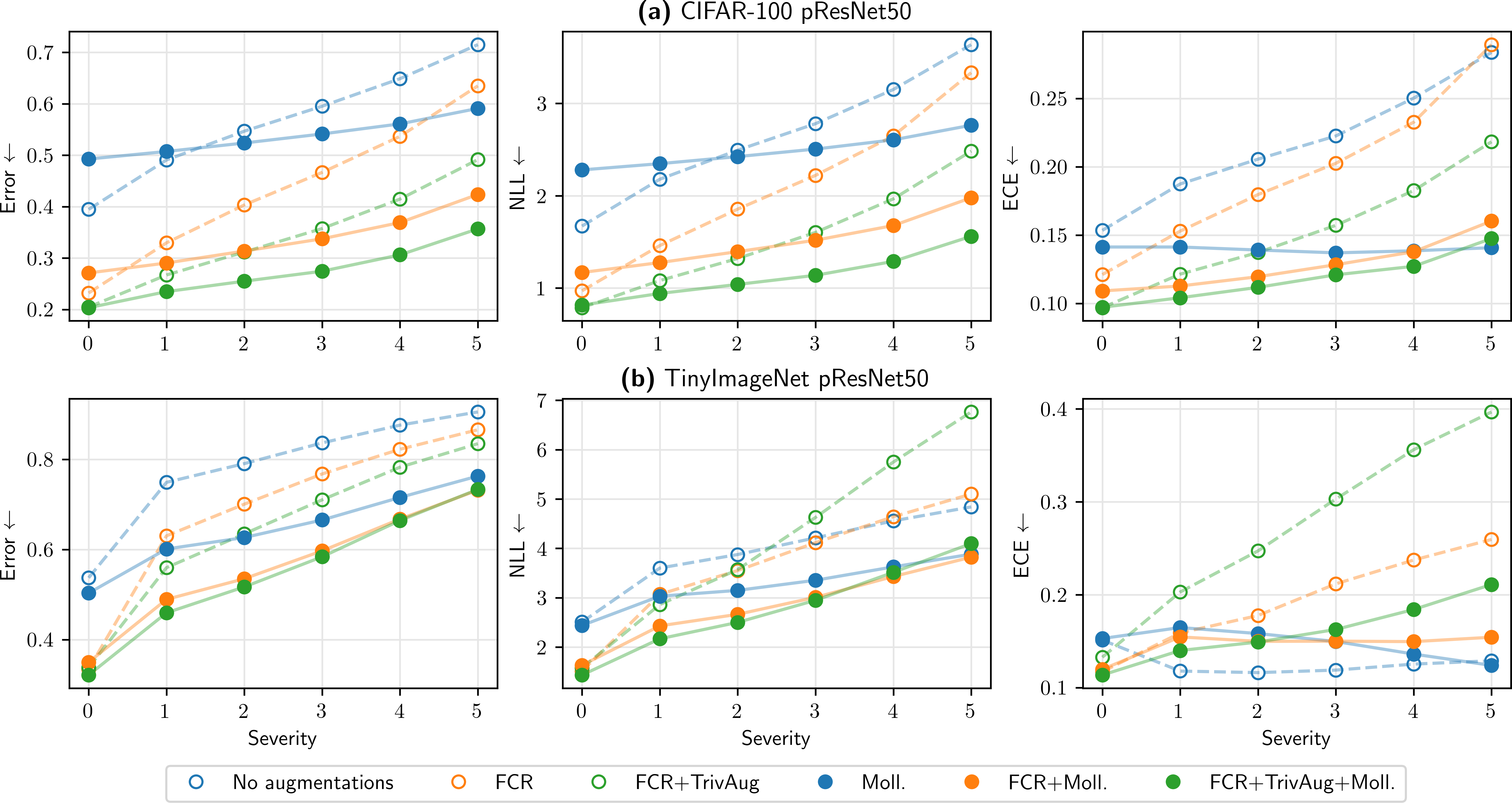}
    \caption{Adding mollification improves augmented models over corruption severity.} 
    \label{fig:severity_appendix}
\end{figure*}

In \cref{fig:severity_appendix} we provide a breakdown of performance metrics for the corrupted CIFAR-100 and TinyImageNet over corruption severity (zero being clean validation data).
It is not surprising that the metrics degrade with intensity of the corruptions.
Meanwhile, it is interesting to observe that mollification attenuates this effect less with performance more stable across corruption severities.

\section{SPECTRAL ANALYSIS}
\label{sec:spectral_analysis}
   
\begin{figure}[h]
    \centering
     \subfloat[][]{\includegraphics[width=0.78\linewidth]{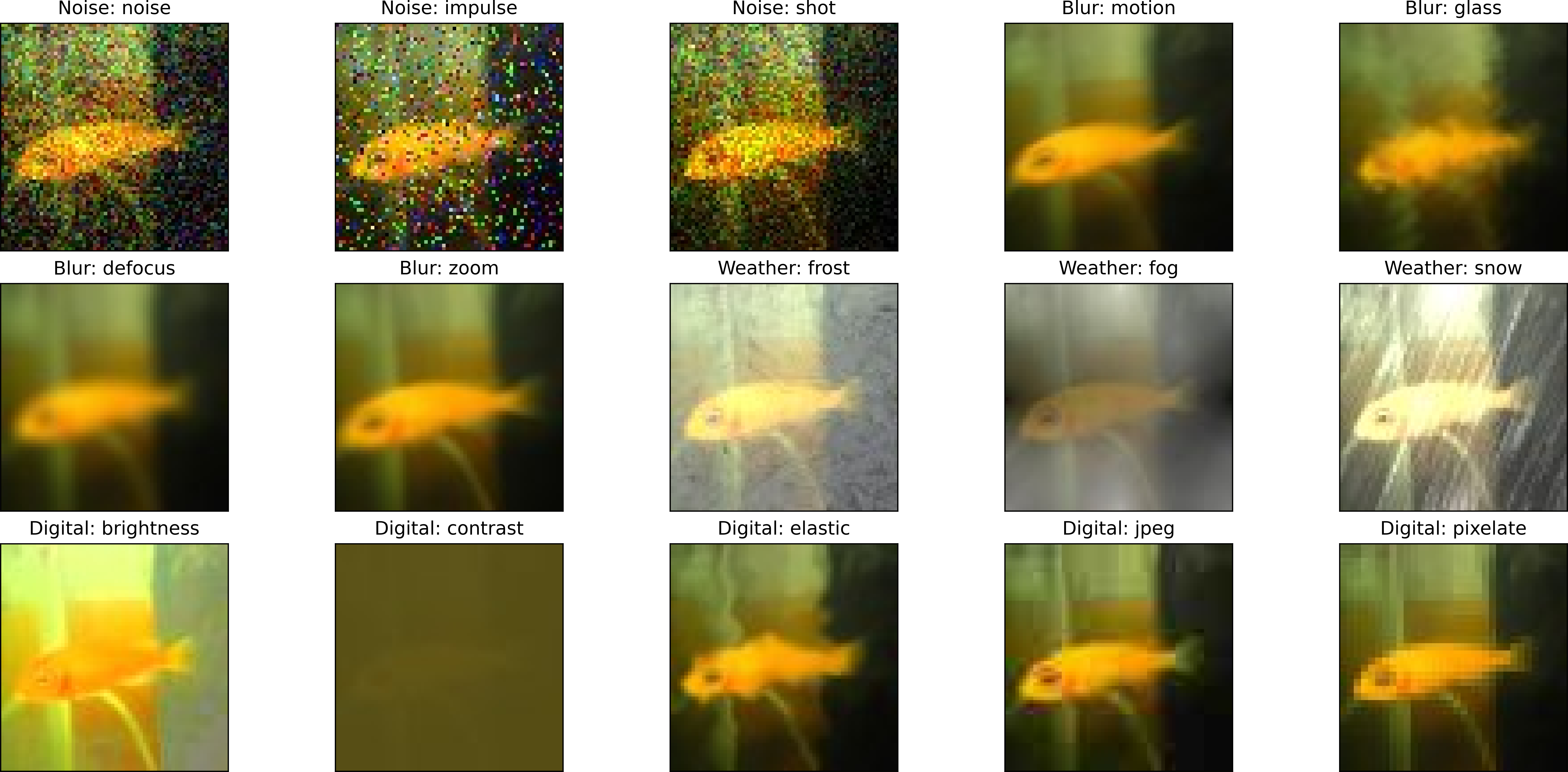}\label{afsa}}
     \\
     \subfloat[][]{\includegraphics[width=0.81\linewidth]{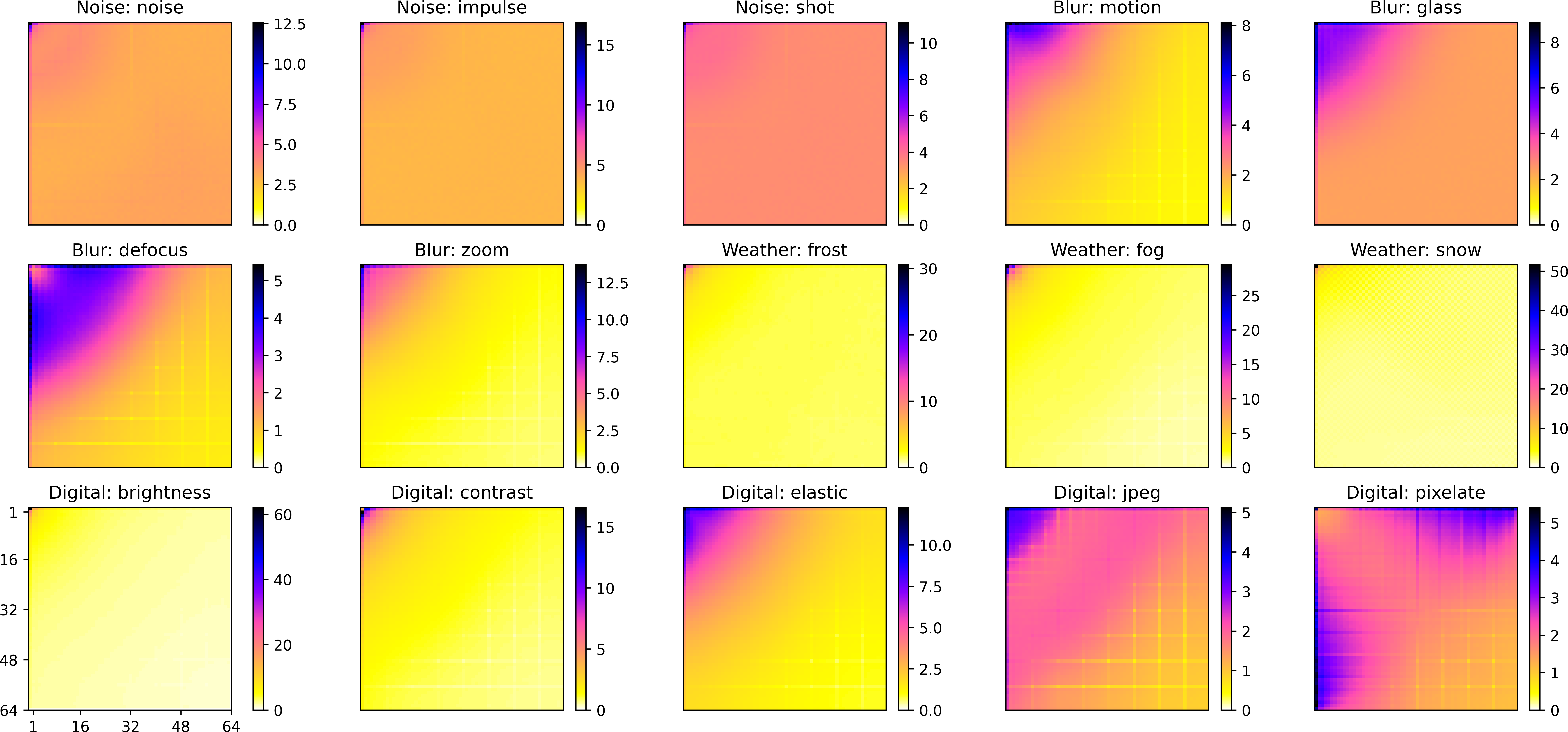}\label{afsa}}
    \caption{The DCT spectral densities \textbf{(b)} of the 15 corruption types \textbf{(a)}. The weather and digital corruptions align resemble the uniform or exponential densities of noise and blur, respectively.}
    \label{fig:density}
\end{figure}

In \cref{fig:density} we compute the average discrete cosine transform (DCT) change from corruptions over TinyImageNet validation set ($N=10,000$ images $\mathbf{x}_i$),
\begin{align}
    \frac{1}{N} \sum_{i=1}^N \operatorname{abs}( \operatorname{DCT}[\operatorname{corruption} [\mathbf{x}_i]] - \operatorname{DCT}[\mathbf{x}_i] )
\end{align}
\cref{fig:density} shows which frequencies are modulated by each corruption type. Noise corruptions affect all frequencies uniformly, while blur corruptions follow an exponential decay on the frequencies. Weather corruptions affect dominantly the low frequencies (top-left) in blur-like pattern. The set of digital corruptions are divided into noise-like and blur-like frequency modulation, with pixelate being an outlier. This analysis suggests spectral similarities between noise/blur and digital/weather corruptions.

\clearpage

\section{NORMALIZING THE CROSS-ENTROPY-BASED LIKELIHOOD}
\label{app:deriv:normalized:likelihood}

Let's consider label smoothing with soft labels defined as
$$
    \y^{\mathrm{LS}} = (1-a) \y^{\mathrm{onehot}} + \frac{a}{C} \mathbf{1},
$$
where we have introduced $a = \gamma_t - 1$. 

We are interested in deriving an expression for a proper cross-entropy-based likelihood function expressing $p(\y^{\mathrm{LS}} | \x; \theta)$.
In other words, we require that $p(\y^{\mathrm{LS}} | \x; \theta)$ is indeed a proper distribution over $\y^{\mathrm{LS}}$.
The difficulty is that the domain of the labels is now continuous and the cross-entropy loss does not correspond to the negative of the logarithm of a properly normalized likelihood. 

Soft labels $\y^{\mathrm{LS}}$ depend on $a$ and $\y^{\mathrm{onehot}}$. 
For simplicity, let's denote the one-hot vectors $\y^{\mathrm{onehot}}$ with a one in position $j$ as $\e_j$.
With this definition, the normalization constant to obtain a proper likelihood can be expressed as follows:
$$
Z = \sum_{j=1}^C \int_{0}^1 \prod_{i=1}^C  f_i^{(\y^{\mathrm{LS}}(a, \e_j))_i}  da \, .
$$
Note that $f_i$ here represent probabilities of class label $i$, that is these are the values post-softmax transformation of the output of the network. 
Without loss of generality, let's focus on the case $j=1$:
$$
\int_{0}^1 \prod_{i=1}^C f_i^{(\y^{\mathrm{LS}}(a, \e_1))_i}  da
=
\int_{0}^1 f_1^{1 - a + \frac{a}{C}} f_2^{\frac{a}{C}} \ldots f_C^{\frac{a}{C}}  da  \, ,
$$
which we can simplify into:
$$
f_1 \int_{0}^1 
f_1^{a\frac{1-C}{C}} f_2^{\frac{a}{C}} \ldots f_C^{\frac{a}{C}}  da
=
f_1 \int_{0}^1 
\left(f_1^{\frac{1-C}{C}} f_2^{\frac{1}{C}} \ldots f_C^{\frac{1}{C}}\right)^{a}  da  \, .
$$

We can proceed in a similar way for any $\e_j$, and we can compact the expression of the integral further as:
$$
f_j \int_{0}^1 
\left( f_j^{-1} \prod_{i=1}^C f_i^{\frac{1}{C}} \right)^{a}  da
=
f_j \int_{0}^1 
\left( \frac{K}{f_j} \right)^{a}  da  \, ,
$$
where we have introduced $K = \prod_{i=1}^C f_i^{\frac{1}{C}}$.

The normalization constant $Z$ is the sum over $j$ of these integrals, for which the solution is in the form:
$$
\int_0^1 z^a \, da = \frac{z - 1}{\log z} \, ,
$$
which yields:
$$
Z = \sum_{j=1}^C f_j \int_0^1 
\left( \frac{K}{f_j} \right)^{a}  da
=
\sum_{j=1}^C \frac{K - f_j}{\log(K) - \log(f_j)}  \, .
$$
This expression requires some care in the implementation to avoid underflows in extreme cases where one of the $f_j$ is close to one, and in practice it is better to attempt computing $\log Z$ instead. 
We tested the cross-entropy-based likelihood with this normalization, but we did not obtain any improvement in performance while complicating the implementation. 

\section{UNBIASED ESTIMATE OF THE MARGINAL AUGMENTED LIKELIHOOD}
\label{sec:inference}

In the main paper, we introduced the marginalized form of the augmented likelihood under transformations $T_\phi(\x)$ of the inputs:
\begin{equation} \label{eq:auglik_app}
    \mathcal{L} = \log p(\D|\theta) = %
    \sum_{n=1}^N \log \int p(\y_n | \x_n, \phi, \theta) p(\phi) d\phi,
\end{equation}
The log-likelihood in \cref{eq:auglik_app} is intractable due to an integral over the space of continuous corruptions or augmentations. Furthermore, simple Monte Carlo averaging of the likelihood within the logarithm is biased for $K > 1$ \citep{durbin1997},
\begin{align}
    \mathcal{L} = \log p(\D|\theta) &= \sum_{n=1}^N \log \int p(\y_n | \x_n, \phi ; \theta) p(\phi) d\phi  \\
    &\approx  \sum_{n=1}^N \log \frac{1}{K} \sum_{k=1}^K p(\y_n | \x_n, \phi_k ; \theta) \, ,
\end{align}
where $\phi_k \sim p(\phi)$. In practice this approach tends to underestimate $\mathcal{L}$.

\paragraph{Lower bound on the log-expectation}
The intractable log-likelihood $\mathcal{L}$ can be approximated in multiple ways. We can simply estimate the biased Jensen posterior by moving the logarithm inside the integral \citep{wenzel20posterior}:
\begin{align}
    \log p(\D|\theta) &= \sum_{n=1}^N \log \int p(\y_n | \x_n, \phi ; \theta) p(\phi) d\phi \label{eq:auglik2} \\
    &\ge \sum_{n=1}^N \int \log p(\y_n | \x_n, \phi_k ; \theta) p(\phi) d\phi \label{eq:lb} \\
    &\approx \sum_{n=1}^N \frac{1}{K} \sum_{k=1}^K \log p(\y_n | \x_n, \phi_k ; \theta), 
\end{align}
where $\phi_k \sim p(\phi)$. This represents a lower bound of the true augmented likelihood, for which Monte-Carlo approximation is unbiased. The Jensen bound can be applied with importance sampling to tighten the bound \citep{burda2015importance,luo2020sumo}.

\paragraph{Bias correction}
We can also apply a bias-correction \citep{durbin1997,shephard1997} to the sample mean of $\exp{(\mathcal{L})}$. %
First, we imagine wanting to estimate the logarithm of an integral $I$, considering a Monte Carlo approximation of the integral $I$ itself,
\begin{align}
    I &= \int p(\y | \x, \phi, \theta) p(\phi) d\phi \\
    I &\approx I_K = \frac{1}{K} \sum_k p(\y | \phi_k, \x, \theta), \quad \phi_k \sim p(\phi).
\end{align}
It can be shown that the estimator $I_K$ can be turn into an unbiased estimator as follows \citep{durbin1997,shephard1997} 
\begin{align}
    \log I &\approx \log I_K + \frac{1}{2}\frac{\var[I_K]}{I_K^2} \\
    \var[I_K] &= \frac{1}{K(K-1)} \sum_k ( p(\y|\x,\phi_k,\theta) - I_K )^2 \\
    0 &= \E\left[\log I_K + \frac{1}{2}\frac{\var[I_K]}{I_K^2}\right].
\end{align}
The correction makes the log-integral unbiased with respect to different samplings $(\phi_1, \ldots, \phi_K) \sim p(\phi)$. With $K=1$ the approximation the log-integral is already unbiased, but it suffers from large variance. In image classification, $K$ refers to the number of corruptions per image within a mini-batch. The correction is most beneficial in small-$K$ regimes. For instance, in Batch Augmentation multiple instances of each image are used within a batch \citep{hoffer2020augment}.

\paragraph{Tempering vs augmented likelihood}

Our marginalized additive augmented likelihood \cref{eq:auglik_app} aligns with the `Jensen' likelihoods discussed by \citet{wenzel20posterior}. In contrast, \citet{kapoor2022uncertainty} argues for multiplicative tempered augmented likelihood 
\begin{align} \label{eq:geomlik}
    p_\mathrm{geom}(\D|\theta) &= \prod_{n=1}^N \prod_{k=1}^K p\big(\y_n | \x_n, \phi, \theta\big)^{1/K},
\end{align}
where the augmentations are aggregated by a geometric mean. These approaches are connected: the geometric likelihood \cref{eq:geomlik} is the lower bound \cref{eq:lb} of the augmented likelihood \cref{eq:auglik2}. In the geometric likelihood we can interpret the augmentations as down-weighted data points, which are assumed to factorize despite not being i.i.d.. In the augmented likelihood we interpret the data points as i.i.d. samples, and augmentations as expanding each sample into a distribution, which leads to a product over observations and integral over augmentations.

\end{document}